\documentclass[pmlr,twocolumn,10pt]{jmlr} 

\mlhtrack{proceedings}

\newif\iffinal
\finaltrue  

\iffinal
    \ifmlhneedspmlr
      \jmlrvolume{346}
      \jmlryear{2026}
    \fi
    \ifmlhfindings \jmlrproceedings{}{ML4H 2026 - Findings Track}\fi
    \ifmlhdemo     \jmlrproceedings{}{ML4H 2026 - Demo Track}\fi
    \jmlrworkshop{Machine Learning for Health (ML4H) 2026}
\else
    \jmlrproceedings{}{Submitted to ML4H 2026: \mlhtrackname}
    \jmlrworkshop{Machine Learning for Health (ML4H) 2026}
\fi

\usepackage[switch]{lineno}

\usepackage{booktabs}
\usepackage{multirow}
\usepackage{listings}
\usepackage{makecell}
\usepackage{enumitem}
\usepackage{xurl}

\colorlet{punct}{red!60!black}
\definecolor{background}{RGB}{245, 245, 245}
\definecolor{delim}{RGB}{20,105,176}
\colorlet{numb}{magenta!60!black}

\lstdefinelanguage{json}{
    basicstyle=\scriptsize\rmfamily,
    numbers=left,
    numberstyle=\scriptsize,
    stepnumber=1,
    numbersep=8pt,
    xleftmargin=16pt,
    showstringspaces=false,
    breaklines=true,
    frame=lines,
    backgroundcolor=\color{background},
    literate=
     *{0}{{{\color{numb}0}}}{1}
      {1}{{{\color{numb}1}}}{1}
      {2}{{{\color{numb}2}}}{1}
      {3}{{{\color{numb}3}}}{1}
      {4}{{{\color{numb}4}}}{1}
      {5}{{{\color{numb}5}}}{1}
      {6}{{{\color{numb}6}}}{1}
      {7}{{{\color{numb}7}}}{1}
      {8}{{{\color{numb}8}}}{1}
      {9}{{{\color{numb}9}}}{1}
      {:}{{{\color{punct}{:}}}}{1}
      {,}{{{\color{punct}{,}}}}{1}
      {\{}{{{\color{delim}{\{}}}}{1}
      {\}}{{{\color{delim}{\}}}}}{1}
      {[}{{{\color{delim}{[}}}}{1}
      {]}{{{\color{delim}{]}}}}{1},
}

\theorembodyfont{\upshape}
\theoremheaderfont{\scshape}
\theorempostheader{:}
\theoremsep{\newline}

\title[Enhancing LLM Medical Coding with Structured External Knowledge]{Enhancing LLM Medical Coding with Structured External Knowledge}

\author{
 \Name{Yidong Gan\nametag{$^{1,2,3}$\thanks{Work done during an internship at Oracle Health and AI.}}} \Email{yidong.gan@sydney.edu.au}\\
 \Name{David D. Nguyen\nametag{$^{1}$}} \Email{david.n.nguyen@oracle.com}\\
 \Name{Yang Lin\nametag{$^{1}$}} \Email{yang.y.lin@oracle.com}\\
 \Name{Peter Zhong\nametag{$^{1}$}} \Email{peter.zhong@oracle.com}\\
 \Name{Thanh Vu\nametag{$^{1}$}} \Email{thanh.v.vu@oracle.com}\\
 \Name{Long Duong\nametag{$^{1}$}} \Email{long.duong@oracle.com}\\
 \Name{Yuan-Fang Li\nametag{$^{1}$}} \Email{yuanfang.li@oracle.com}\\
 \addr $^{1}$Oracle Health and AI \quad $^{2}$The University of Sydney \quad $^{3}$CSIRO
}

\begin{document}

\maketitle

\ifmlhdemo\else
\begin{abstract}
Accurate medical coding requires consulting authoritative resources such as the ICD tabular list and coding guidelines. Existing LLM-based automated methods largely rely on internal model knowledge, which is prone to hallucination and cannot keep pace with guideline updates. We introduce RAG-Coding, an agentic, training-free method that augments LLMs with structured external knowledge. Specifically, we encode the tabular list as a knowledge graph capturing hierarchical and instructional code relationships, and distil the guidelines into concise, code-specific summaries rather than retrieving raw text. On the MDACE dataset, RAG-Coding outperforms the strongest LLM-based baseline by 3--13\% in micro-F1 and 1--8\% in macro-F1 across five LLM backbones; compared to the supervised state-of-the-art, PLM-ICD, RAG-Coding achieves 11\% higher recall at the cost of 6\% lower precision. We also release MDACE-2025, an expert re-annotation of the MDACE test set under the contemporary 2025 ICD-10-CM/PCS guidelines. On MDACE-2025, RAG-Coding also outperforms all baselines notably in micro- and macro-F1, demonstrating effective generalisation to guideline updates. Ablation studies show stepwise gains, confirming the value of structured external knowledge for LLM-based medical coding.
\end{abstract}
\begin{keywords}
Medical NLP, RAG, Medical coding, ICD-10-CM/PCS, Data Resources
\end{keywords}
\fi

\ifmlhneedsstatements
\paragraph*{Data and Code Availability}
Our experiments use the MDACE dataset~\citep{cheng-etal-2023-mdace}. We additionally construct and release a re-annotated subset of MDACE for this study. The original MDACE dataset is available under the access conditions provided by its authors. Due to intellectual property constraints, the source code for this study is not released.

\paragraph*{Institutional Review Board (IRB)}
Our study does not require IRB approval.
\fi

\section{Introduction}
\begin{figure*}[t]
    \centering
    \includegraphics[width=\linewidth]{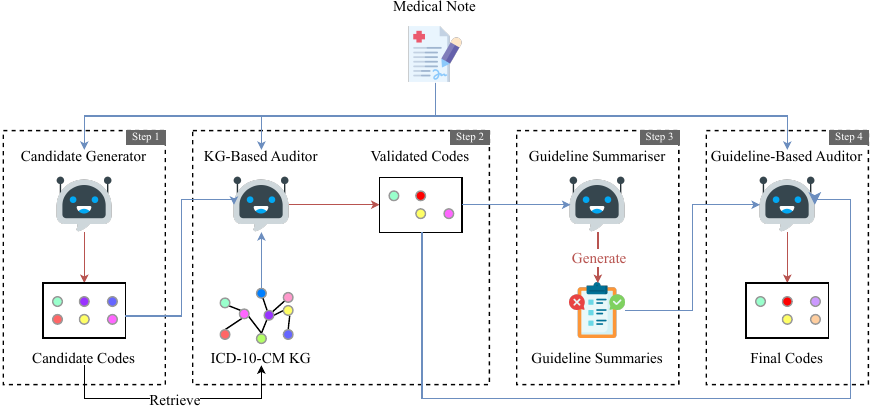}
    \caption{Overall architecture of the proposed RAG-Coding method. Blue lines denote input to the agents. Red lines denote output of the agents. Black lines denote intermediary operations that do not involve any agents.}
    \label{fig:ragcoding_architecture}
\end{figure*}
Medical coding maps clinical documentation to standardised alphanumeric codes, forming the foundation of hospital billing, medical research, and national health policy~\citep{dong2022automated}. Traditional manual coding is time-consuming, and some hospitals experience backlogs extending over a year~\citep{alonso2020problems}. This burden has motivated research into automated medical coding~\citep{edin2023automated}.

Recent advancements in large language models (LLMs) have prompted research into their application for medical coding, leveraging their instruction-following abilities and extensive parametric knowledge. However, early results \citep{boyle2023automated, falis2024can} suggest LLMs, whether open-source or commercial, are not yet reliable coders \emph{out-of-the-box}, demonstrating no performance advantage over PLM-ICD~\citep{huang-etal-2022-plm, edin2023automated}, the BERT-based supervised state-of-the-art (SOTA).

Accurate medical coding requires understanding authoritative coding resources, which professional coders routinely consult~\citep{dong2022automated, motzfeldt-etal-2025-code}. Two primary examples are the ICD-10-CM tabular list and the ICD-10-CM coding guidelines (see \S~\ref{sec:preliminaries}). However, existing LLM-based automated methods largely overlook them, relying solely on LLMs' internal knowledge, which is prone to hallucination and cannot keep pace with the annual updates of coding resources~\citep{guidelines-update-freq}. Existing datasets have similarly failed to keep pace: widely used datasets such as MIMIC-III, MIMIC-IV, and MDACE~\citep{johnson2016mimic, johnson2023mimic, cheng-etal-2023-mdace} were annotated under pre-2022 guidelines, limiting evaluation against current clinical standards. MIMIC's annotations are also noisy and lack justification for coding decisions~\citep{cheng-etal-2023-mdace}, while MDACE omits code sequencing---a mandatory workflow component that distinguishes principal from secondary diagnoses and is required for reimbursement.

To address the methodological gap, we introduce \textbf{RAG-Coding}, an agentic, training-free automated coding method. How external knowledge is represented matters as much as whether it is available to the LLM: RAG-Coding grounds LLM coding decisions in structured knowledge, encoding the ICD-10-CM tabular list as a knowledge graph (KG) that captures hierarchical and instructional code relationships, and distilling the coding guidelines into concise, code-specific summaries rather than retrieving them as raw text. RAG-Coding decomposes the coding task into four sequential steps: (1) generate candidate codes; (2) validate them using the KG; (3) retrieve and summarise relevant guidelines into code-specific rules; and (4) further validate using those summaries (see Figure~\ref{fig:ragcoding_architecture}).

Additionally, to support evaluation against current clinical standards, we provide \textbf{MDACE-2025}, expert re-annotations of the MDACE test set under the 2025 coding guidelines, with code sequencing and justification comments for complex coding decisions. Such coding justifications are important as they make annotated codes auditable, clinically traceable, and easier for human coders to verify and correct.

Our contributions are as follows:
\begin{enumerate}[noitemsep]
    \item We develop RAG-Coding, an agentic, training-free method supporting the full U.S.\ ICD-10-CM system ($\approx$70K codes), grounding LLM decisions in a KG built from the tabular list and code-specific guideline summaries.
    \item We release MDACE-2025, the only public ICD coding dataset with 2025-guideline annotations, code sequencing, and expert justifications, opening research on guideline drift, principal-diagnosis coding, and error analysis.
    \item RAG-Coding achieves strong results on both MDACE and MDACE-2025, demonstrating effective generalisation to guideline shifts and showing that the structural representation of external knowledge matters as much as the knowledge itself for LLM-based medical coding.
\end{enumerate}

\section{Related Work}

\subsection{Automated Medical Coding}
Automated medical coding has been studied for decades, from rule-based systems~\citep{farkas2008automatic} to multi-label classification~\citep{mullenbach-etal-2018-explainable, cao-etal-2020-hypercore} and BERT-based models pre-trained on biomedical corpora, such as PLM-ICD~\citep{huang-etal-2022-plm}, the current supervised SOTA~\citep{edin2023automated, gan-etal-2025-aligning, motzfeldt-etal-2025-code}.

LLMs have been applied to ICD coding in zero- and few-shot settings~\citep{yang2023multi, boyle2023automated}, but remain prone to hallucination and fail to adhere to mandatory coding rules. Retrieval-augmented generation (RAG)~\citep{lewis2020retrieval} has emerged as a practical solution~\citep{baksi-etal-2025-medcoder, motzfeldt-etal-2025-code}. To our knowledge, CLH~\citep{motzfeldt-etal-2025-code}  is the only existing baseline that uses coding guidelines, differing from RAG-Coding in how knowledge is integrated: plain instructional notes and guideline text vs.\ a KG and code-specific summaries (example in Appendix~\ref{appx:clh_vs_ragcoding}).

\subsection{Structured Knowledge Integration}
Prior work has integrated structured knowledge in neural networks for medical coding: \citet{rios-kavuluru-2018-shot} encode ICD parent-child relationships via a graph convolutional network over label embeddings, enabling generalisation to rare and unseen codes; related efforts incorporate code synonyms~\citep{yuan-etal-2022-code} and co-occurrence graphs~\citep{cao-etal-2020-hypercore}. RAG-Coding extends this line of work by integrating structured ICD knowledge with LLMs at inference time, rather than during training.

\subsection{Medical Coding Datasets} \label{sec:related_work_datasets}
MIMIC-III, MIMIC-IV, and MDACE are widely used public ICD datasets~\citep{johnson2016mimic, johnson2023mimic, cheng-etal-2023-mdace}, though MIMIC's annotations are noisy and lack justification for coding decisions~\citep{dong2022automated, cheng-etal-2023-mdace}. MDACE addresses these quality concerns via professionally re-annotated ICD-10-CM/PCS codes for a subset of MIMIC-III, but all three datasets were annotated under pre-2022 guidelines, and MDACE omits code sequencing---a mandatory component of real-world coding workflows. These limitations motivate our introduction of MDACE-2025, described in \S~\ref{sec:mdace2025}.

\section{Background: Common Coding Knowledge Sources} \label{sec:preliminaries}
Below are two authoritative medical coding resources commonly used by human coders:

\paragraph{ICD-10-CM Tabular List.}
The tabular list groups codes in a tree-structured hierarchy, from broad categories to specific diagnoses, and contains instructional notes that define code usage and relationships, such as: \textit{includes} (scope of a broad code category), \textit{inclusion\_term} (synonymous or included terms), \textit{use\_additional\_code} (codes that must also be assigned), and \textit{excludes1} (mutually exclusive codes).

\paragraph{ICD-10-CM Coding Guidelines.}
The guidelines span over 100 pages of narrative rules that coders follow, comprising general conventions (overall coding principles) and chapter-specific instructions (nuances within particular disease categories). A snippet of the guidelines' table of contents is available in Figure~\ref{fig:guidelines_retrieval}.

\section{MDACE-2025} \label{sec:mdace2025}
We introduce MDACE-2025, providing expert re-annotations of the MDACE test set under the April 2025 ICD-10-CM/PCS coding guidelines. This allows us to validate RAG-Coding against both the original MDACE ground truth (2021 guidelines) and current clinical standards. Because RAG-Coding and the evaluated LLM-based baselines are \textbf{training-free}, we re-annotated only the test set, adding code sequencing and expert justifications for complex cases.

\paragraph{Annotation Process.}
Re-annotation was conducted by a professional coder holding the AHIMA Certified Coding Specialist (CCS) and AAPC Certified Professional Coder (CPC) credentials; CCS is a mastery-level credential focused on inpatient coding, MDACE's exact setting. Rather than annotating each encounter from scratch, the coder used a \emph{Review and Revise} procedure: starting from the original MDACE annotations as a baseline, the coder systematically reviewed and revised each encounter for conformance with the 2025 guidelines, correcting codes affected by guideline changes or coding errors. This procedure, combined with the coder's expertise, likely reduces annotation errors due to misinterpretation. However, fully disentangling error sources would require a more controlled, multi-annotator study, which is beyond the scope of this work.

\paragraph{Annotation Shift.}
We quantify the shift between MDACE and MDACE-2025 using two rates: \emph{retention}, the fraction of original MDACE codes that persist in MDACE-2025, and \emph{addition}, the fraction of MDACE-2025 codes that are new relative to MDACE. These rates are computed over code instances (i.e., each code counted once per encounter in which it is assigned), not unique codes; unique code counts are reported separately in Table~\ref{tab:stats_diff_between_datasets}. Overall, MDACE-2025 retains 527/678 (77.73\%) of MDACE's codes and adds 378/905 (41.77\%) new codes. Diagnosis (ICD-10-CM) codes are comparatively stable, largely reflecting expanded coverage of Z and R codes, whereas procedure (ICD-10-PCS) codes shift far more, consistent with the 2025 guidelines' stricter requirement that procedure codes be explicitly supported by operative notes. Full per-category statistics and distribution shifts are in Appendix~\ref{appx:annotation_diff}.

\paragraph{Research Value.}
MDACE-2025 enables research directions not supported by existing public datasets: evaluating generalisation to evolving guidelines, targeted study of principal diagnosis coding (mandatory in real-world reimbursement workflows) via code sequencing, and error analysis via expert justifications.

\section{Method}
We introduce RAG-Coding, which consists of four main steps: (1) initial ICD code assignment, (2) KG-based validation, (3) guideline retrieval and summarisation, and (4) final validation using guideline summaries, each executed by one agent. Any predicted codes that do not appear in the ICD-10-CM Tabular List (i.e., hallucinated codes) are removed deterministically. Our methodological core lies in steps 2 and 3, which define \emph{how} external coding knowledge is structurally encoded. Agent prompts are in Appendix~\ref{appx:prompts}.

\subsection{Problem Definition}
Let $x_i$ denote an input medical note; the objective is to map $x_i$ to $y_i$, a set of valid ICD codes. This task is traditionally framed as multi-label classification, where models predict over a fixed set of codes. Following recent studies~\citep{yang2023multi, boyle2023automated, baksi-etal-2025-medcoder, motzfeldt-etal-2025-code}, we instead frame it as a generation problem: given an LLM-based system $\mathcal{S}$, $y_i = \mathcal{S}(x_i)$. This lets $\mathcal{S}$ produce any ICD code, whereas classification-based approaches are limited to codes seen during training.

\subsection{Step 1: Initial ICD Code Assignment} \label{sec:rag_coding_step1}
This step narrows the candidate pool to an initial set of relevant ICD codes. Prior studies~\citep{baksi-etal-2025-medcoder, motzfeldt-etal-2025-code} achieve this via named entity recognition (NER) combined with semantic search---extracting diagnoses from medical notes and matching them to ICD code descriptions---achieving strong recall but weak precision. \citet{baksi-etal-2025-medcoder} show that out-of-the-box LLMs tasked with direct ICD code generation face the same precision challenge despite similar recall. Building on this, we use dynamic two-shot examples (see Appendix~\ref{appx:few_shot}) and instruct our \textbf{\textit{Candidate Generator}} agent to assign relevant ICD codes and supporting evidence simultaneously, strengthening the link between documented diagnoses and their codes.

Although the generated evidence could, as in the NER $+$ semantic search approach, be used to retrieve further codes, we choose not to: our preliminary finding shows that the generated code set largely overlaps with the retrieved code set, offering marginal recall gains at a notable precision cost.

\subsection{Step 2: Knowledge Graph Retrieval and Validation}
This step validates the candidate codes (output of step 1) with a KG that captures their relationships. We construct the KG once, by parsing the official ICD-10-CM tabular list, and reuse it for all medical notes. We denote the complete KG as $G = (V, E)$, where $V$ is the set of nodes, consisting of all ICD-10-CM codes and instructional notes (e.g., code synonyms); $E$ is the set of edges, indicating the hierarchical and instructional properties between the nodes. For example, [I12.9 $\to$ \textit{ancestor} $\to$ I12] and [I12.9 $\to$ \textit{inclusion\_term} $\to$ ``Hypertensive chronic kidney disease NOS, Hypertensive renal disease NOS''].

Validation begins by retrieving a subgraph $G_S=(V_S, E_S) \subseteq G$, where $V_S$ contains the candidate codes, their ancestors, and associated instructional notes, and $E_S$ contains the edges connected to $V_S$. Our \textbf{\textit{KG-Based Auditor}} agent then validates the candidates using the rich context in $G_S$, resolving issues such as incorrect codes (e.g., codes whose descriptions contradict the clinical note), conflicting codes (e.g., codes that mutually exclude each other), and missing codes (e.g., complementary codes that must be assigned together).

\subsection{Step 3: Guideline Retrieval and Summarisation}
This step retrieves and summarises relevant coding guidelines for the validated codes (output of step 2). Given a code, our \textbf{\textit{Guideline Summariser}} agent is prompted with its textual description and the guidelines' table of contents to locate the applicable guideline sections. For example, for code I27.20 (Pulmonary hypertension, unspecified), the agent retrieves \S I.A.9.b (``Unspecified'' Codes), \S I.B.18 (Use of Sign/Symptom/Unspecified Codes), and \S I.C.9 (Diseases of the Circulatory System). An illustration of this retrieval process is shown in Figure~\ref{fig:guidelines_retrieval}.

Although using the table of contents to retrieve chapter-specific guidelines may appear redundant, it ensures consideration of all relevant rules, as a code may relate to others in different chapters. For example, elevated blood pressure (R03.0) can indicate underlying hypertension (I10--I16), and guidelines in Section I.C.9 (for codes I00--I99) explicitly prohibit coding R03.0 when hypertension is definitively diagnosed. Since multiple sections may be retrieved, and some span over 10 pages, filtering is essential: our \textbf{\textit{Guideline Summariser}} agent condenses them into bullet points containing only the rules applicable to the code being validated (see Table~\ref{tab:clh_vs_ragcoding_example} for an example).

\subsection{Step 4: Final Validation Using Guideline Summaries}
This step further verifies the validated codes (output of step 2) using guideline summaries (output of step 3), ensuring consistency with $G_S$ and clinical guidelines. Our \textbf{\textit{Guideline-Based Auditor}} agent cross-references the input medical note and guideline summaries, deciding to retain, remove, or replace each code with an alternative. When no guideline summary is available for a code, the default action is to retain it, as it was derived from $G_S$ and no official guideline evidence supports removing or replacing it.

\section{Experimental Setup}
\subsection{Dataset}
We evaluate RAG-Coding on MDACE and MDACE-2025, which are introduced in Sections~\ref{sec:related_work_datasets} and~\ref{sec:mdace2025}, respectively. We choose MDACE for three reasons: (1) paired with MDACE-2025, it allows us to assess generalisation to evolving clinical standards; (2) it provides high-quality, high-agreement annotations; and (3) it offers standardised splits, ensuring comparability and reproducibility. While both datasets contain ICD-10-PCS annotations, we evaluate only on ICD-10-CM codes, matching the LLM-based baselines' exclusive ICD-10-CM focus.

\subsection{Metrics}
Each encounter consists of one or more notes (e.g., discharge summary, radiology report), which we process independently during inference. During evaluation, predictions and gold-standard codes are merged at the encounter level: for note $x_i$ with gold-standard codes $y_i$ and predicted codes $\hat{y}_i$, we aggregate across all notes in the same patient encounter $c_j$ as $y_j = \bigcup_{x_i \in c_j} y_i$ and $\hat{y}_j = \bigcup_{x_i \in c_j} \hat{y}_i$. This \emph{encounter-level} evaluation reflects real-world practice~\citep{cheng-etal-2023-mdace}, in which coders review all notes from an encounter when assigning codes\footnote{We followed the same practice when annotating MDACE-2025.}. Following prior work~\citep{boyle2023automated, li2024exploring, baksi-etal-2025-medcoder, motzfeldt-etal-2025-code}, we report micro- and macro-averaged precision, recall, and F1 on the test set. Micro-averaging aggregates counts across all code occurrences; macro-averaging computes the metric separately per code and averages equally.

\subsection{Implementation Details} \label{sec:implementation}
We experiment with five different LLM backbones for our RAG-Coding method: Deepseek-V3.2~\citep{liu2025deepseek}, Qwen3~\citep{yang2025qwen3}, GPT-4o~\citep{openai2024gpt4o}, GPT-4.1~\citep{openai2025gpt4dot1}, and GPT-5 with low reasoning mode~\citep{openai2025gpt5}. All LLMs are hosted locally or privately via the Azure OpenAI service to follow PhysioNet's data use agreement\footnote{\url{https://physionet.org/news/post/gpt-responsible-use}}. Exact LLM checkpoints are detailed in Appendix~\ref{appx:llm_checkpoints_and_guideline_versions}.

\subsection{Baselines} \label{sec:baselines}
Our RAG-Coding method is compared against Tree Search~\citep{boyle2023automated}, MedCodER~\citep{baksi-etal-2025-medcoder}, CLH~\citep{motzfeldt-etal-2025-code}, and PLM-ICD~\citep{huang-etal-2022-plm}. Full descriptions of the baselines are in Appendix~\ref{appx:description_of_baselines}. All LLM-based baselines use the same setup (e.g., same backbones, hyperparameters, and knowledge sources), except dynamic few-shot retrieval, which is specific to RAG-Coding's step 1 and not directly applicable to the baselines (justified in Appendix~\ref{appx:description_of_baselines}); we ablate its effect in \S\ref{sec:ablation_study}. PLM-ICD is the SOTA supervised baseline, trained on ICD-10-CM codes from MIMIC-IV and the MDACE training set ($\approx$16K unique codes, covering $\approx$20\% of the full ICD-10-CM code space).

\begin{table*}[t!]
\centering
\footnotesize
\renewcommand{\arraystretch}{0.9}
\begin{tabular}{c|l|l*{6}{c}}
\toprule
\#C & LLM & Method & \multicolumn{3}{c}{Micro} & \multicolumn{3}{c}{Macro} \\
\cmidrule(lr){4-6}
\cmidrule(lr){7-9}
 & & & Precision & Recall & F1 & Precision & Recall & F1 \\
\midrule
\multirow{20}{*}{70K} & \multirow{4}{*}{Qwen3}
& Tree Search & 0.08 & 0.17 & 0.11 & 0.05 & 0.05 & 0.04 \\
& & MedCodER & 0.29 & 0.36 & 0.32 & 0.16 & 0.16 & 0.15 \\
& & CLH & 0.27 & \textbf{0.49} & 0.35 & 0.17 & \textbf{0.18} & 0.17 \\
& & RAG-Coding  & \textbf{0.43} & 0.39 & \textbf{0.41} & \textbf{0.20} & \textbf{0.18} & \textbf{0.18} \\
\cmidrule{2-9}
& \multirow{4}{*}{Deepseek-V3.2}
& Tree Search & 0.12 & 0.25 & 0.16 & 0.07 & 0.08 & 0.07 \\
& & MedCodER & 0.33 & \textbf{0.48} & 0.39 & 0.18 & 0.19 & 0.17 \\
& & CLH & 0.37 & 0.46 & 0.41 & 0.21 & 0.21 & 0.21 \\
& & RAG-Coding  & \textbf{0.46} & 0.43 & \textbf{0.44} & \textbf{0.25} & \textbf{0.22} & \textbf{0.22} \\
\cmidrule{2-9}
& \multirow{4}{*}{GPT-4o}
& Tree Search & 0.14 & 0.26 & 0.18 & 0.09 & 0.09 & 0.09 \\
& & MedCodER & 0.39 & 0.36 & 0.37 & 0.19 & 0.18 & 0.17 \\
& & CLH & 0.29 & 0.47 & 0.36 & 0.18 & 0.19 & 0.18 \\
& & RAG-Coding  & \textbf{0.42} & \textbf{0.48} & \textbf{0.45} & \textbf{0.21} & \textbf{0.22} & \textbf{0.21} \\
\cmidrule{2-9}
& \multirow{4}{*}{GPT-4.1}
& Tree Search & 0.11 & 0.22 & 0.15 & 0.08 & 0.08 & 0.07 \\
& & MedCodER & 0.31 & 0.44 & 0.36 & 0.17 & 0.18 & 0.17 \\
& & CLH & 0.32 & 0.52 & 0.40 & 0.20 & 0.22 & 0.20 \\
& & RAG-Coding  & \textbf{0.47} & \textbf{0.59} & \textbf{0.53} & \textbf{0.27} & \textbf{0.29} & \textbf{0.27} \\
\cmidrule{2-9}
& \multirow{4}{*}{GPT-5}
& Tree Search & 0.17 & 0.33 & 0.22 & 0.10 & 0.11 & 0.09 \\
& & MedCodER & 0.37 & 0.42 & 0.40 & 0.21 & 0.22 & 0.20 \\
& & CLH & 0.40 & 0.55 & 0.46 & 0.24 & 0.26 & 0.24 \\
& & RAG-Coding  & \textbf{0.46} & \textbf{0.66} & \textbf{0.54} & \textbf{0.31} & \textbf{0.33} & \textbf{0.32} \\
\midrule
\multirow{5}{*}{16K} & \multirow{5}{*}{GPT-5}
& PLM-ICD & \textbf{0.54} & 0.53 & 0.53 & 0.29 & 0.30 & 0.30 \\
& & Tree Search & 0.19 & 0.33 & 0.24 & 0.11 & 0.12 & 0.10 \\
& & MedCodER & 0.39 & 0.42 & 0.41 & 0.23 & 0.23 & 0.21 \\
& & CLH & 0.43 & 0.55 & 0.48 & 0.26 & 0.29 & 0.27 \\
& & RAG-Coding  & 0.48 & \textbf{0.64} & \textbf{0.55} & \textbf{0.30} & \textbf{0.32} & \textbf{0.31} \\
\bottomrule
\end{tabular}
\renewcommand{\arraystretch}{1}
\caption{
Results on the \emph{MDACE} test set. ``\#C'' is the number of candidate codes. Full \#C=16K results with other LLM backbones are in Appendix~\ref{appx:full_results}.
}
\label{tab:main_results}
\end{table*}

\section{Results}
The main result values reported in this section represent means over three runs; standard deviations (typically $\le$ 0.03) are available in Appendix~\ref{appx:full_results}.

\subsection{Main Results on MDACE}
The \#C=70K results in Table~\ref{tab:main_results} compare RAG-Coding against LLM-based baselines on the MDACE test set. Using uniform LLM backbones, RAG-Coding outperforms the best baseline by 3--13\% in micro-F1 and 1--8\% in macro-F1. With GPT-5, RAG-Coding achieves a micro-F1 of 0.54 and macro-F1 of 0.32, outperforming the best baseline (CLH) by 8\% in both micro- and macro-F1 over the full ICD-10-CM code space ($\approx$70K codes). These results confirm that structured in-domain knowledge enhances LLMs' medical coding performance.

PLM-ICD is a supervised, classification-based method that can only predict codes seen during training. We trained PLM-ICD with the $\approx$16K codes available in the MIMIC-IV and MDACE training set; comparing it directly against LLM-based methods operating over the full 70K code space would be unfair to PLM-ICD. We therefore post hoc filter all LLM-based methods' outputs, including RAG-Coding's, to this 16K code space and present the corresponding \#C=16K results in Table~\ref{tab:main_results}. While not a perfectly apples-to-apples comparison---all of the LLM-based methods still reason over the larger space initially---the filter narrows the inference space gap. Under this setting, RAG-Coding achieves higher micro recall than PLM-ICD ($+11\%$), while PLM-ICD achieves higher micro precision ($+6\%$). Overall, RAG-Coding marginally outperforms PLM-ICD in both micro- and macro-F1. Following \citet{motzfeldt-etal-2025-code}, we attribute PLM-ICD's strength to its supervised training on data from the same ICU environment as the MDACE test set, which the training-free LLM-based methods do not benefit from.

\subsection{Main Results on MDACE-2025}
\begin{table*}[t!]
\centering
\footnotesize
\begin{tabular}{c|l|l*{6}{c}}
\toprule
\#C & LLM & Method & \multicolumn{3}{c}{Micro} & \multicolumn{3}{c}{Macro} \\
\cmidrule(lr){4-6}
\cmidrule(lr){7-9}
 & & & Precision & Recall & F1 & Precision & Recall & F1 \\
\midrule
\multirow{4}{*}{70K} & \multirow{4}{*}{GPT-5}
& Tree Search & 0.20 & 0.28 & 0.24 & 0.12 & 0.13 & 0.12 \\
& & MedCodER & 0.43 & 0.35 & 0.38 & 0.24 & 0.23 & 0.23 \\
& & CLH & 0.45 & 0.44 & 0.45 & 0.26 & 0.27 & 0.26 \\
& & RAG-Coding & \textbf{0.57} & \textbf{0.55} & \textbf{0.56} & \textbf{0.34} & \textbf{0.34} & \textbf{0.33} \\
\midrule
\multirow{5}{*}{16K} & \multirow{5}{*}{GPT-5}
& PLM-ICD & \textbf{0.62} & 0.39 & 0.48 & 0.27 & 0.25 & 0.26 \\
& & Tree Search & 0.23 & 0.28 & 0.25 & 0.13 & 0.13 & 0.11 \\
& & MedCodER & 0.46 & 0.34 & 0.40 & 0.25 & 0.24 & 0.23 \\
& & CLH & 0.49 & 0.42 & 0.45 & 0.27 & 0.28 & 0.26 \\
& & RAG-Coding & 0.58 & \textbf{0.55} & \textbf{0.56} & \textbf{0.35} & \textbf{0.34} & \textbf{0.34} \\
\bottomrule
\end{tabular}
\caption{
Results on the \emph{MDACE-2025} test set. ``\#C'' is the number of candidate codes; all LLM-based methods use GPT-5. Full results with other backbones are in Appendix~\ref{appx:full_results}.
}
\label{tab:results_against_baselines_mdace2025}
\end{table*}

The main difference between MDACE and MDACE-2025 is the latter's expanded code annotations, especially for R (symptom-related) and Z (health status-related) codes (see Appendix~\ref{appx:annotation_diff}). Table~\ref{tab:results_against_baselines_mdace2025} compares RAG-Coding against baselines on MDACE-2025: relative to MDACE, all methods show higher precision but proportionally lower recall, yielding comparable F1 across datasets. The precision gain suggests both RAG-Coding and the baselines correctly identify newly added codes previously penalised as false positives; however, the expanded annotations also make high recall harder.

Table~\ref{tab:results_against_baselines_mdace2025} shows that RAG-Coding, with GPT-5, outperforms PLM-ICD by 8\% in both micro- and macro-F1. As noted in prior work~\citep{edin2023automated}, PLM-ICD can struggle to distinguish between similar Z codes under class imbalance. This limitation is more visible in MDACE-2025, which provides more comprehensive Z-code annotations. Moreover, the evaluated PLM-ICD model was trained on older MIMIC-IV and MDACE data, reducing its ability to adapt to ground-truth shifts introduced by updated coding guidelines. In contrast, RAG-Coding uses the same up-to-date tabular code list and guidelines used to annotate MDACE-2025, yielding better conformance and generalisation to its ground truth.

\subsection{Ablation Study and Analysis} \label{sec:ablation_study}
Unless otherwise noted, all ablations in this section are single runs under the \#C=70K setting (GPT-4.1 backbone). Table~\ref{tab:ablation_step_by_step} shows stepwise gains, confirming the contribution of each RAG-Coding component. Our central claim, however, is not merely that external knowledge helps, but that its structural \emph{representation} matters as much as its availability. We test this directly with two symmetric ablations that hold the underlying knowledge content fixed while varying only its format. First, at step 2, we replace our KG tabular list with CLH-style plain text, holding step 1 fixed and reporting scores after step 2. As shown in Table~\ref{tab:ablation_guideline_format} (top), this reduces micro-F1 from 0.47 to 0.41, a 6-point drop, despite identical underlying tabular list content. Separately, at step 3, we replace our guideline summaries with CLH-style narrative snippets while holding steps 1, 2, and 4 fixed and reporting scores after step 4. As shown in Table~\ref{tab:ablation_guideline_format} (bottom), this reduces micro-F1 by 7 points despite identical upstream outputs and downstream validation. Although these ablations do not assess summary faithfulness, they show that the structural format of external knowledge is a key driver of RAG-Coding's gains, helping explain its 3--13\% micro-F1 advantage over CLH across five backbones (Table~\ref{tab:main_results}). To verify this margin is not confounded by RAG-Coding's dynamic few-shot in step 1, we also run the full RAG-Coding pipeline at $k=0$: micro-F1 drops 5 points (0.51 $\to$ 0.46) but still exceeds CLH's full pipeline (0.40) by 6 points (Table~\ref{tab:main_results}).

To further disentangle the role of external knowledge, we evaluate a closed-book variant of RAG-Coding. After generating candidate ICD codes in step 1, the same LLM validates them using only its parametric knowledge. As detailed in Appendix~\ref{appx:self_correction}, the results show that LLMs cannot effectively self-correct coding errors without external knowledge.

\begin{table}[t!]
\centering
\footnotesize
\begin{tabular}{l*{3}{c}}
\toprule
 & \multicolumn{3}{c}{Micro} \\
\cmidrule(lr){2-4}
& Precision & Recall & F1 \\
\midrule
RAG-Coding (step 1) & 0.34 & \textbf{0.54} & 0.41 \\
RAG-Coding (steps 1--2) & 0.42 & 0.53 & 0.47 \\
RAG-Coding (steps 1--4) & \textbf{0.48} & 0.54 & \textbf{0.51} \\
\bottomrule
\end{tabular}
\caption{
Ablation study on the \emph{MDACE} test set. Step 3 outputs guideline summaries, which cannot be directly evaluated by classification metrics. Therefore, results for steps 1--3 are omitted in this table.
}
\label{tab:ablation_step_by_step}
\end{table}

\begin{table}[t!]
\centering
\footnotesize
\begin{tabular}{ll*{3}{c}}
\toprule
Component & Format & \multicolumn{3}{c}{Micro} \\
\cmidrule(lr){3-5}
 & & Precision & Recall & F1 \\
\midrule
\multirow{2}{*}{\makecell[c]{Tabular\\List}} & KG & \textbf{0.42} & \textbf{0.53} & \textbf{0.47} \\
 & Plain-text & 0.38 & 0.44 & 0.41 \\
\midrule
\multirow{2}{*}{\makecell[c]{Coding\\Guidelines}} & Structural & \textbf{0.48} & \textbf{0.54} & \textbf{0.51} \\
 & Narrative & 0.45 & 0.43 & 0.44 \\
\bottomrule
\end{tabular}
\caption{
Ablations isolating the effect of knowledge \emph{representation} on the \emph{MDACE} test set.
}
\label{tab:ablation_guideline_format}
\end{table}

We also manually examined RAG-Coding's intermediate outputs for 20 random test samples, categorising changes made by each agent (GPT-4.1 backbone). We summarise the key findings below; the full per-component analysis is in Appendix~\ref{appx:detailed_analysis}.

\textbf{\textit{Candidate Generator (step 1)}} achieves moderate recall (0.41) but low precision (0.27). False positives concentrate in symptom and injury codes; false negatives concentrate in Z-codes and circulatory codes.

\textbf{\textit{KG-Based Auditor (step 2)}} removes 2.2 codes per sample on average, with 91\% removal accuracy (39 of 43). It correctly removes codes whose descriptions contradict the medical note (e.g., A41.53 \emph{Sepsis due to Serratia} when the note documents Klebsiella) and symptom codes rendered redundant by more specific diagnoses. Its primary failure is removing briefly mentioned but valid comorbidities, where the KG provides no conflicting signal, suggesting the removal is likely due to insufficient contextual density.

\textbf{\textit{Guideline Summariser (step 3)}} retrieves summaries for 90\% (115 of 128) of codes. Gaps are mainly in circulatory and respiratory codes, where rules are defined only at the chapter level rather than per code. When no guideline is found, the code is retained by default in step 4.

\textbf{\textit{Guideline-Based Auditor (step 4)}} achieves 89\% removal accuracy (17 of 19). Code replacements are its most valuable corrections (e.g., I10 \emph{Hypertension} replaced by I12.9 \emph{Hypertensive chronic kidney disease} when chronic kidney disease is co-documented). However, addition accuracy is poor (2 of 10): the agent over-interprets ``prefer specificity'' guidelines, adding specific codes without sufficient supporting evidence in the note.

\section{Conclusion and Future Work}
We propose RAG-Coding, an agentic, training-free method for automated medical coding that grounds LLM decisions in structurally encoded external knowledge, yielding substantial precision and F1 gains over recent LLM-based baselines. Compared to PLM-ICD, RAG-Coding consistently achieves higher recall and could be generalised to the full ICD-10-CM code space. Ablations confirm that both the KG and guideline summaries are critical to these gains, showing that how external knowledge is structurally represented matters as much as its availability to the LLM. We also release MDACE-2025, extending MDACE with 2025-guideline annotations, code sequencing, and justification comments; RAG-Coding also outperforms all baselines on this dataset, showing effective generalisation to updated guidelines. In future work, we plan to expand MDACE-2025's annotation scope, re-annotating the full MDACE dataset with more annotators and fine-grained labels for each annotation change (e.g., guideline change vs.\ error correction). We also plan to study the faithfulness of the guideline summaries, the impact of dataset–guideline mismatch on RAG-Coding, and severity-weighted metrics using MDACE-2025's code sequencing, since some errors (e.g., principal diagnosis errors) carry greater clinical and financial consequences than others.

\section*{Limitations}
Most recent LLM-based medical coding methods (e.g., our evaluated baselines) focus on the English language and the ICD-10-CM ontology. Our evaluation therefore focuses on the same setting. While extending RAG-Coding and the baselines to other languages and ontologies is valuable, it requires non-trivial effort and would fit better as a separate paper.

As a multi-stage pipeline, RAG-Coding is prone to error propagation: a code missed upstream is difficult (though not impossible; see Appendix~\ref{appx:detailed_analysis}) to recover downstream. Our manual analysis shows recall errors concentrate in Z-codes and circulatory codes, suggesting a targeted second-pass candidate-generation step as future work.

MDACE-2025 was re-annotated by a single credentialed coder rather than multiple independent annotators; the coder's dual CCS/CPC credentials and Review-and-Revise procedure (see \S\ref{sec:mdace2025}) mitigate but do not eliminate misinterpretation risk, and disentangling guideline-driven shifts from annotator bias would need a larger, multi-annotator study.

\bibliography{custom}

@inproceedings{edin2023automated,
  title={Automated medical coding on {MIMIC-III} and {MIMIC-IV}: A critical review and replicability study},
  author={Edin, Joakim and Junge, Alexander and Havtorn, Jakob D and Borgholt, Lasse and Maistro, Maria and Ruotsalo, Tuukka and Maal{\o}e, Lars},
  booktitle={Proceedings of the 46th International ACM SIGIR Conference on Research and Development in Information Retrieval},
  pages={2572--2582},
  year={2023}
}

@article{johnson2016mimic,
  title={{MIMIC-III}, a freely accessible critical care database},
  author={Johnson, Alistair EW and Pollard, Tom J and Shen, Lu and Lehman, Li-wei H and Feng, Mengling and Ghassemi, Mohammad and Moody, Benjamin and Szolovits, Peter and Anthony Celi, Leo and Mark, Roger G},
  journal={Scientific Data},
  volume={3},
  number={1},
  pages={1--9},
  year={2016},
  publisher={Nature Publishing Group}
}

@article{johnson2023mimic,
  title={{MIMIC-IV}, a freely accessible electronic health record dataset},
  author={Johnson, Alistair EW and Bulgarelli, Lucas and Shen, Lu and Gayles, Alvin and Shammout, Ayad and Horng, Steven and Pollard, Tom J and Hao, Sicheng and Moody, Benjamin and Gow, Brian and others},
  journal={Scientific Data},
  volume={10},
  number={1},
  pages={1},
  year={2023},
  publisher={Nature Publishing Group UK London}
}

@article{dong2022automated,
  title={Automated clinical coding: what, why, and where we are?},
  author={Dong, Hang and Falis, Mat{\'u}{\v{s}} and Whiteley, William and Alex, Beatrice and Matterson, Joshua and Ji, Shaoxiong and Chen, Jiaoyan and Wu, Honghan},
  journal={npj Digital Medicine},
  volume={5},
  number={1},
  pages={159},
  year={2022},
  publisher={Nature Publishing Group UK London}
}

@inproceedings{kim2021read,
  title={Read, attend, and code: Pushing the limits of medical codes prediction from clinical notes by machines},
  author={Kim, Byung-Hak and Ganapathi, Varun},
  booktitle={Machine Learning for Healthcare Conference},
  pages={196--208},
  year={2021},
  organization={PMLR}
}

@article{alonso2020problems,
  title={Problems and barriers during the process of clinical coding: a focus group study of coders’ perceptions},
  author={Alonso, Vera and Santos, Jo{\~a}o Vasco and Pinto, Marta and Ferreira, Joana and Lema, Isabel and Lopes, Fernando and Freitas, Alberto},
  journal={Journal of Medical Systems},
  volume={44},
  pages={1--8},
  year={2020},
  publisher={Springer}
}

@article{gao2024optimising,
  title={Optimising the paradigms of human {AI} collaborative clinical coding},
  author={Gao, Yue and Chen, Yuepeng and Wang, Minghao and Wu, Jinge and Kim, Yunsoo and Zhou, Kaiyin and Li, Miao and Liu, Xien and Fu, Xiangling and Wu, Ji and others},
  journal={npj Digital Medicine},
  volume={7},
  number={1},
  pages={368},
  year={2024},
  publisher={Nature Publishing Group UK London}
}

@misc{openai2024gpt4o,
   author = "OpenAI",
   title = "Hello {GPT-4o}",
   year = "2024",
   howpublished = {\url{https://openai.com/index/hello-gpt-4o}},
   note = "[Online; accessed 1-Dec-2025]"
 }

@misc{openai2025gpt4dot1,
   author = "OpenAI",
   title = "Introducing {GPT-4.1} in the {API}",
   year = "2025",
   howpublished = {\url{https://openai.com/index/gpt-4-1/}},
   note = "[Online; accessed 1-Dec-2025]"
 }

@misc{openai2025gpt5,
   author = "OpenAI",
   title = "{GPT-5} is here",
   year = "2025",
   howpublished = {\url{https://openai.com/gpt-5}},
   note = "[Online; accessed 12-Oct-2025]"
 }

@inproceedings{
    boyle2023automated,
    title={Automated clinical coding using off-the-shelf large language models},
    author={Joseph Boyle and Antanas Kascenas and Pat Lok and Maria Liakata and Alison O'Neil},
    booktitle={Deep Generative Models for Health Workshop NeurIPS 2023},
    year={2023},
    url={https://openreview.net/forum?id=mqnR8rGWkn}
}

@article{falis2024can,
  title={Can {GPT-3.5} generate and code discharge summaries?},
  author={Falis, Mat{\'u}{\v{s}} and Gema, Aryo Pradipta and Dong, Hang and Daines, Luke and Basetti, Siddharth and Holder, Michael and Penfold, Rose S and Birch, Alexandra and Alex, Beatrice},
  journal={Journal of the American Medical Informatics Association},
  volume={31},
  number={10},
  pages={2284--2293},
  year={2024},
  publisher={Oxford Academic}
}

@inproceedings{motzfeldt-etal-2025-code,
    title = "Code Like Humans: A Multi-Agent Solution for Medical Coding",
    author = "Motzfeldt, Andreas Geert  and
      Edin, Joakim  and
      Christensen, Casper L.  and
      Hardmeier, Christian  and
      Maal{\o}e, Lars  and
      Rogers, Anna",
    editor = "Christodoulopoulos, Christos  and
      Chakraborty, Tanmoy  and
      Rose, Carolyn  and
      Peng, Violet",
    booktitle = "Findings of the Association for Computational Linguistics: EMNLP 2025",
    month = nov,
    year = "2025",
    address = "Suzhou, China",
    publisher = "Association for Computational Linguistics",
    url = "https://aclanthology.org/2025.findings-emnlp.1231/",
    pages = "22612--22627",
    ISBN = "979-8-89176-335-7"
}

@article{li2024exploring,
  title={Exploring {LLM} multi-agents for {ICD} coding},
  author={Li, Rumeng and Wang, Xun and Yu, Hong},
  journal={arXiv preprint arXiv:2406.15363},
  year={2024}
}

@article{farkas2008automatic,
  title={Automatic construction of rule-based {ICD-9-CM} coding systems},
  author={Farkas, Rich{\'a}rd and Szarvas, Gy{\"o}rgy},
  journal={BMC Bioinformatics},
  volume={9},
  number={Suppl 3},
  pages={S10},
  year={2008},
  publisher={Springer}
}

@article{madaan2023self,
  title={Self-refine: Iterative refinement with self-feedback},
  author={Madaan, Aman and Tandon, Niket and Gupta, Prakhar and Hallinan, Skyler and Gao, Luyu and Wiegreffe, Sarah and Alon, Uri and Dziri, Nouha and Prabhumoye, Shrimai and Yang, Yiming and others},
  journal={Advances in Neural Information Processing Systems},
  volume={36},
  pages={46534--46594},
  year={2023}
}

@article{lewis2020retrieval,
  title={Retrieval-augmented generation for knowledge-intensive {NLP} tasks},
  author={Lewis, Patrick and Perez, Ethan and Piktus, Aleksandra and Petroni, Fabio and Karpukhin, Vladimir and Goyal, Naman and K{\"u}ttler, Heinrich and Lewis, Mike and Yih, Wen-tau and Rockt{\"a}schel, Tim and others},
  journal={Advances in Neural Information Processing Systems},
  volume={33},
  pages={9459--9474},
  year={2020}
}

@inproceedings{yang2023multi,
  title={Multi-label few-shot {ICD} coding as autoregressive generation with prompt},
  author={Yang, Zhichao and Kwon, Sunjae and Yao, Zonghai and Yu, Hong},
  booktitle={Proceedings of the AAAI Conference on Artificial Intelligence},
  volume={37},
  pages={5366--5374},
  year={2023}
}

@article{douze2025faiss,
  title={The {Faiss} library},
  author={Douze, Matthijs and Guzhva, Alexandr and Deng, Chengqi and Johnson, Jeff and Szilvasy, Gergely and Mazar{\'e}, Pierre-Emmanuel and Lomeli, Maria and Hosseini, Lucas and J{\'e}gou, Herv{\'e}},
  journal={IEEE Transactions on Big Data},
  year={2025},
  publisher={IEEE}
}

@misc{embedding2025openai,
   author = "OpenAI",
   title = "text-embedding-3-large",
   year = "2025",
   howpublished = {\url{https://platform.openai.com/docs/models/text-embedding-3-large}},
   note = "[Online; accessed 1-Dec-2025]"
}

@article{liu2025deepseek,
  title={{DeepSeek-V3.2}: Pushing the frontier of open large language models},
  author={Liu, Aixin and Mei, Aoxue and Lin, Bangcai and Xue, Bing and Wang, Bingxuan and Xu, Bingzheng and Wu, Bochao and Zhang, Bowei and Lin, Chaofan and Dong, Chen and others},
  journal={arXiv preprint arXiv:2512.02556},
  year={2025}
}

@article{yang2025qwen3,
  title={Qwen3 technical report},
  author={Yang, An and Li, Anfeng and Yang, Baosong and Zhang, Beichen and Hui, Binyuan and Zheng, Bo and Yu, Bowen and Gao, Chang and Huang, Chengen and Lv, Chenxu and others},
  journal={arXiv preprint arXiv:2505.09388},
  year={2025}
}

@inproceedings{rios-kavuluru-2018-shot,
    title = "Few-Shot and Zero-Shot Multi-Label Learning for Structured Label Spaces",
    author = "Rios, Anthony  and
      Kavuluru, Ramakanth",
    editor = "Riloff, Ellen  and
      Chiang, David  and
      Hockenmaier, Julia  and
      Tsujii, Jun{'}ichi",
    booktitle = "Proceedings of the 2018 Conference on Empirical Methods in Natural Language Processing",
    month = oct # "-" # nov,
    year = "2018",
    address = "Brussels, Belgium",
    publisher = "Association for Computational Linguistics",
    url = "https://aclanthology.org/D18-1352/",
    doi = "10.18653/v1/D18-1352",
    pages = "3132--3142"
}

@misc{guidelines-update-freq,
   author = "CMS",
   title = "{ICD-10} Coordination and Maintenance Committee Meetings",
   year = "2026",
   howpublished = {\url{https://www.cms.gov/medicare/coding-billing/icd-10-codes/icd-10-coordination-maintenance-committee-meetings}},
   note = "[Online; accessed 24-May-2026]"
 }

@inproceedings{devlin-etal-2019-bert,
    title = "{BERT}: Pre-training of Deep Bidirectional Transformers for Language Understanding",
    author = "Devlin, Jacob  and
      Chang, Ming-Wei  and
      Lee, Kenton  and
      Toutanova, Kristina",
    editor = "Burstein, Jill  and
      Doran, Christy  and
      Solorio, Thamar",
    booktitle = "Proceedings of the 2019 Conference of the North {A}merican Chapter of the Association for Computational Linguistics: Human Language Technologies, Volume 1 (Long and Short Papers)",
    month = jun,
    year = "2019",
    address = "Minneapolis, Minnesota",
    publisher = "Association for Computational Linguistics",
    url = "https://aclanthology.org/N19-1423/",
    doi = "10.18653/v1/N19-1423",
    pages = "4171--4186"
}

@inproceedings{gan-etal-2025-aligning,
    title = "Aligning {AI} Research with the Needs of Clinical Coding Workflows: Eight Recommendations Based on {US} Data Analysis and Critical Review",
    author = "Gan, Yidong  and
      Rybinski, Maciej  and
      Hachey, Ben  and
      Kummerfeld, Jonathan K.",
    editor = "Che, Wanxiang  and
      Nabende, Joyce  and
      Shutova, Ekaterina  and
      Pilehvar, Mohammad Taher",
    booktitle = "Proceedings of the 63rd Annual Meeting of the Association for Computational Linguistics (Volume 1: Long Papers)",
    month = jul,
    year = "2025",
    address = "Vienna, Austria",
    publisher = "Association for Computational Linguistics",
    url = "https://aclanthology.org/2025.acl-long.45/",
    doi = "10.18653/v1/2025.acl-long.45",
    pages = "909--922",
    ISBN = "979-8-89176-251-0"
}

@inproceedings{huang-etal-2022-plm,
    title = "{PLM}-{ICD}: Automatic {ICD} Coding with Pretrained Language Models",
    author = "Huang, Chao-Wei  and
      Tsai, Shang-Chi  and
      Chen, Yun-Nung",
    editor = "Naumann, Tristan  and
      Bethard, Steven  and
      Roberts, Kirk  and
      Rumshisky, Anna",
    booktitle = "Proceedings of the 4th Clinical Natural Language Processing Workshop",
    month = jul,
    year = "2022",
    address = "Seattle, WA",
    publisher = "Association for Computational Linguistics",
    url = "https://aclanthology.org/2022.clinicalnlp-1.2/",
    doi = "10.18653/v1/2022.clinicalnlp-1.2",
    pages = "10--20"
}

@inproceedings{baksi-etal-2025-medcoder,
    title = "{M}ed{C}od{ER}: A Generative {AI} Assistant for Medical Coding",
    author = "Baksi, Krishanu Das  and
      Soba, Elijah  and
      Higgins, John J  and
      Saini, Ravi  and
      Wood, Jaden  and
      Cook, Jane  and
      Scott, Jack I  and
      Pudota, Nirmala  and
      Weninger, Tim  and
      Bowen, Edward  and
      Bhattacharya, Sanmitra",
    editor = "Chen, Weizhu  and
      Yang, Yi  and
      Kachuee, Mohammad  and
      Fu, Xue-Yong",
    booktitle = "Proceedings of the 2025 Conference of the Nations of the Americas Chapter of the Association for Computational Linguistics: Human Language Technologies (Volume 3: Industry Track)",
    month = apr,
    year = "2025",
    address = "Albuquerque, New Mexico",
    publisher = "Association for Computational Linguistics",
    url = "https://aclanthology.org/2025.naacl-industry.37/",
    doi = "10.18653/v1/2025.naacl-industry.37",
    pages = "449--459",
    ISBN = "979-8-89176-194-0"
}

@inproceedings{cheng-etal-2023-mdace,
    title = "{MDACE}: {MIMIC} Documents Annotated with Code Evidence",
    author = "Cheng, Hua  and
      Jafari, Rana  and
      Russell, April  and
      Klopfer, Russell  and
      Lu, Edmond  and
      Striner, Benjamin  and
      Gormley, Matthew",
    editor = "Rogers, Anna  and
      Boyd-Graber, Jordan  and
      Okazaki, Naoaki",
    booktitle = "Proceedings of the 61st Annual Meeting of the Association for Computational Linguistics (Volume 1: Long Papers)",
    month = jul,
    year = "2023",
    address = "Toronto, Canada",
    publisher = "Association for Computational Linguistics",
    url = "https://aclanthology.org/2023.acl-long.416/",
    doi = "10.18653/v1/2023.acl-long.416",
    pages = "7534--7550"
}

@inproceedings{cao-etal-2020-hypercore,
    title = "{H}yper{C}ore: Hyperbolic and Co-graph Representation for Automatic {ICD} Coding",
    author = "Cao, Pengfei  and
      Chen, Yubo  and
      Liu, Kang  and
      Zhao, Jun  and
      Liu, Shengping  and
      Chong, Weifeng",
    editor = "Jurafsky, Dan  and
      Chai, Joyce  and
      Schluter, Natalie  and
      Tetreault, Joel",
    booktitle = "Proceedings of the 58th Annual Meeting of the Association for Computational Linguistics",
    month = jul,
    year = "2020",
    address = "Online",
    publisher = "Association for Computational Linguistics",
    url = "https://aclanthology.org/2020.acl-main.282/",
    doi = "10.18653/v1/2020.acl-main.282",
    pages = "3105--3114"
}

@inproceedings{yuan-etal-2022-code,
    title = "Code Synonyms Do Matter: Multiple Synonyms Matching Network for Automatic {ICD} Coding",
    author = "Yuan, Zheng  and
      Tan, Chuanqi  and
      Huang, Songfang",
    editor = "Muresan, Smaranda  and
      Nakov, Preslav  and
      Villavicencio, Aline",
    booktitle = "Proceedings of the 60th Annual Meeting of the Association for Computational Linguistics (Volume 2: Short Papers)",
    month = may,
    year = "2022",
    address = "Dublin, Ireland",
    publisher = "Association for Computational Linguistics",
    url = "https://aclanthology.org/2022.acl-short.91/",
    doi = "10.18653/v1/2022.acl-short.91",
    pages = "808--814"
}

@inproceedings{mullenbach-etal-2018-explainable,
    title = "Explainable Prediction of Medical Codes from Clinical Text",
    author = "Mullenbach, James  and
      Wiegreffe, Sarah  and
      Duke, Jon  and
      Sun, Jimeng  and
      Eisenstein, Jacob",
    editor = "Walker, Marilyn  and
      Ji, Heng  and
      Stent, Amanda",
    booktitle = "Proceedings of the 2018 Conference of the North {A}merican Chapter of the Association for Computational Linguistics: Human Language Technologies, Volume 1 (Long Papers)",
    month = jun,
    year = "2018",
    address = "New Orleans, Louisiana",
    publisher = "Association for Computational Linguistics",
    url = "https://aclanthology.org/N18-1100/",
    doi = "10.18653/v1/N18-1100",
    pages = "1101--1111"
}

@misc{bls2025medicalrecords,
   author = "BLS",
   title = "Medical Records Specialists: Occupational Outlook Handbook",
   year = "2025",
   howpublished = {\url{https://www.bls.gov/ooh/healthcare/medical-records-and-health-information-technicians.htm}},
   note = "[Online; accessed 24-May-2026]"
 }

@article{jin2023medcpt,
  title={{MedCPT}: Contrastive pre-trained transformers with large-scale {PubMed} search logs for zero-shot biomedical information retrieval},
  author={Jin, Qiao and Kim, Won and Chen, Qingyu and Comeau, Donald C and Yeganova, Lana and Wilbur, W John and Lu, Zhiyong},
  journal={Bioinformatics},
  volume={39},
  number={11},
  pages={btad651},
  year={2023},
  publisher={Oxford University Press}
}

@misc{balachandran2024medembed,
  author = {Balachandran, Abhinand},
  title = {{MedEmbed}: Medical-Focused Embedding Models},
  year = {2024},
  url = {https://github.com/abhinand5/MedEmbed}
}

\clearpage

\appendix

\section{Prompts for RAG-Coding Agents} \label{appx:prompts}
Here we present the system prompts used by the agents. The agent's inputs (e.g., the medical note, candidate codes, and $G_S$) are sent as the user message. For the \textbf{\textit{Candidate Generator}}, each of the retrieved examples (Appendix~\ref{appx:few_shot}) is inserted as a user/assistant turn pair.

\subsection{Candidate Generator}
\begin{lstlisting}[language=json]
You are an expert clinical coder.

Instructions:
Your task is to output all relevant ICD-10-CM codes that are relevant to the medical note.
Each code must be supported by one or more verbatim evidence taken directly from the medical note (no paraphrasing, interpretation, or rewriting).
Do not infer or code any condition not explicitly documented. For example, if the note mentions 'smokes 1 pack per day for many years', use F17.20 (nicotine dependence, unspecified), not F17.210 (nicotine dependence, cigarettes), as the type of tobacco product is not clearly documented.

JSON Output Format (you must strictly follow this format and output nothing else):
{
    "results": [
        {
            "code": "<alphanumeric ICD-10-CM code>",
            "description": "<official description of the code>",
            "evidence": [
                "<text snippet that directly supports the code>",
                (continue for all relevant evidence)
            ],
        },
        (continue for all relevant ICD-10-CM codes)
    ]
}
\end{lstlisting}

\subsection{KG-Based Auditor}
\begin{lstlisting}[language=json]
You are an expert medical coder.

Instruction:
Select the most appropriate ICD-10-CM codes from the provided list based on the patient's note.
Your reasoning must be grounded in the provided knowledge graph.

Input Format:
<Note> ... </Note> - The patient's medical note.
<Codes> ... </Codes> - A comma-separated list of potential ICD-10-CM codes.
<KnowledgeGraph> ... </KnowledgeGraph> - A structured list of relationships providing additional context.

Knowledge Graph Format:
The knowledge graph consists of triplets in the format [subject, predicate, object], where:
Subject: An ICD-10-CM code or a range (e.g. I10, I10-I16).
Predicate: The type of relationship, such as:
    "description" - textual explanation of the code.
    "inclusion_term" - synonymous or included terms.
    "use_additional_code" - indicates codes that must also be used.
    "code_first" - indicates the code that should be listed first.
    "ancestor" - links to a parent or higher-level code.
    "excludes1" - codes that must not be used together.
    "includes" - conditions or terms encompassed by the code.
Object: The value or linked entity (e.g. text, Boolean, or another ICD code).

Output Format (you must strictly follow this JSON format and output nothing else):
{
    "results": [
        {
            "code": "<alphanumeric ICD-10-CM code>",
            "justification": "<brief justification for assigning this code>"
        },
        (continue for all selected/assigned ICD-10-CM codes)
    ]
}
\end{lstlisting}

\subsection{Guideline-Based Auditor}
\begin{lstlisting}[language=json]
You are an expert medical coder.

Your task is to audit a provided medical note against a list of assigned ICD-10-CM codes using official coding guidelines. You *must not* use any abbreviations in your response.

**Auditing Process**
For each code, use this format:
Code: [The code to audit]
Guidelines: [Must be fetched via the get_relevant_summaries_for_code() tool]
Thought: [If guidelines are found, reason about whether the code should be retained, removed, or replaced (by an alternative code). If guidelines are not found, **retain the code and move to the next code**]
Decision: [Whether to retain/remove/replace the code]
... (this Code/Guidelines/Thought/Decision should repeat N times, where N is the number of codes)

**Output Format**
Output only the refined codes in this exact format (nothing else):
Final Answer: Code1, Code2, Code3
\end{lstlisting}

Note: The \texttt{get\_relevant\_summaries\_for\_code()} tool takes an ICD-10-CM code and returns its cached guideline summary generated by the \textbf{\textit{Guideline Summariser}} (step 3).

\section{Dynamic Few-Shot for Candidate Generator} \label{appx:few_shot}
We use FAISS~\citep{douze2025faiss} as the vector store. All notes in the MDACE training set are indexed using text-embedding-3-large~\citep{embedding2025openai} as the embedding model. We store the 3072-dimensional, unit-norm embeddings in an exact \texttt{IndexFlatIP} index without quantisation, so retrieval ranks notes by cosine similarity. For each inference instance (medical note), we compute its embedding using the same embedding model and retrieve the top $k$ most similar indexed samples. These retrieved samples serve as in-context examples for the \textbf{\textit{Candidate Generator}} (step 1). As shown in Table~\ref{tab:dynamic_few_shot}, we experimented with $k \in \{0,1,2,3\}$ and found that $k=2$ yields the best F1 score on the MDACE validation set.

A separate ablation on the MDACE validation set (Table~\ref{tab:fewshot_coverage}) shows that, with text-embedding-3-large and $k=2$, the retrieved examples cover only 29\% of the gold codes, and 59\% of the correct codes predicted in step 1 (with GPT-4.1 as the backbone) do not appear among them. Medical embedding models such as MedCPT~\citep{jin2023medcpt} and MedEmbed~\citep{balachandran2024medembed} do not improve retrieval. Step 1 therefore relies substantially on the LLM's parametric knowledge.

\begin{table}[t!]
\centering
\footnotesize
\begin{tabular}{l*{3}{c}}
\toprule
 & \multicolumn{3}{c}{Micro} \\
\cmidrule(lr){2-4}
Few-Shot Size & Precision & Recall & F1 \\
\midrule
$k=0$ & 0.30 & \textbf{0.53} & 0.38 \\
$k=1$ & 0.36 & 0.51 & 0.42 \\
$k=2$ & \textbf{0.40} & 0.51 & \textbf{0.44} \\
$k=3$ & 0.38 & 0.50 & 0.43 \\
\bottomrule
\end{tabular}
\caption{
Results of the \textbf{\textit{Candidate Generator}} agent using different numbers of dynamic few-shot examples on the \emph{MDACE} validation set.
}
\label{tab:dynamic_few_shot}
\end{table}

\begin{table}[t!]
\centering
\footnotesize
\setlength{\tabcolsep}{4pt}
\begin{tabular}{lc}
\toprule
Candidate Source & Gold Codes Covered \\
\midrule
MDACE training set (upper bound) & 0.78 (490/625) \\
\midrule
\multicolumn{2}{l}{\textit{Top-2 retrieved examples, by embedding model}} \\
text-embedding-3-large & \textbf{0.29} (183/625) \\
MedCPT & 0.26 (161/625) \\
MedEmbed & 0.25 (159/625) \\
\bottomrule
\end{tabular}
\caption{
Coverage of gold ICD-10-CM codes by few-shot examples on the \emph{MDACE} validation set. A gold code counts as covered if it is assigned to at least one example for that encounter. The first row uses all \emph{MDACE} training encounters as examples, an upper bound for retrieval since retrieved examples are drawn from the training set. The remaining rows use the top-2 examples retrieved by each embedding model. MedEmbed and MedCPT accept at most 512 tokens, so we average the embeddings of 512-token chunks.
}
\label{tab:fewshot_coverage}
\end{table}

\section{Retrieving Coding Guidelines}
\begin{figure}[t]
    \centering
    \includegraphics[width=\linewidth]{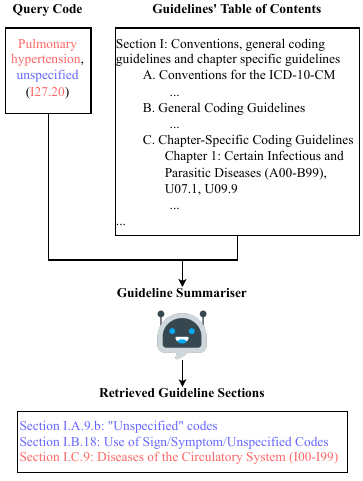}
    \caption{
    Illustration of our \textbf{\textit{Guideline Summariser}} agent's retrieval process for an example code.
    }
    \label{fig:guidelines_retrieval}
\end{figure}

Figure~\ref{fig:guidelines_retrieval} illustrates our \textbf{\textit{Guideline Summariser}} agent's retrieval process for an example code I27.20 (pulmonary hypertension, unspecified).

\section{Comparison Between MDACE and MDACE-2025} \label{appx:annotation_diff}
\begin{figure}[t]
    \centering
    \includegraphics[width=\linewidth]{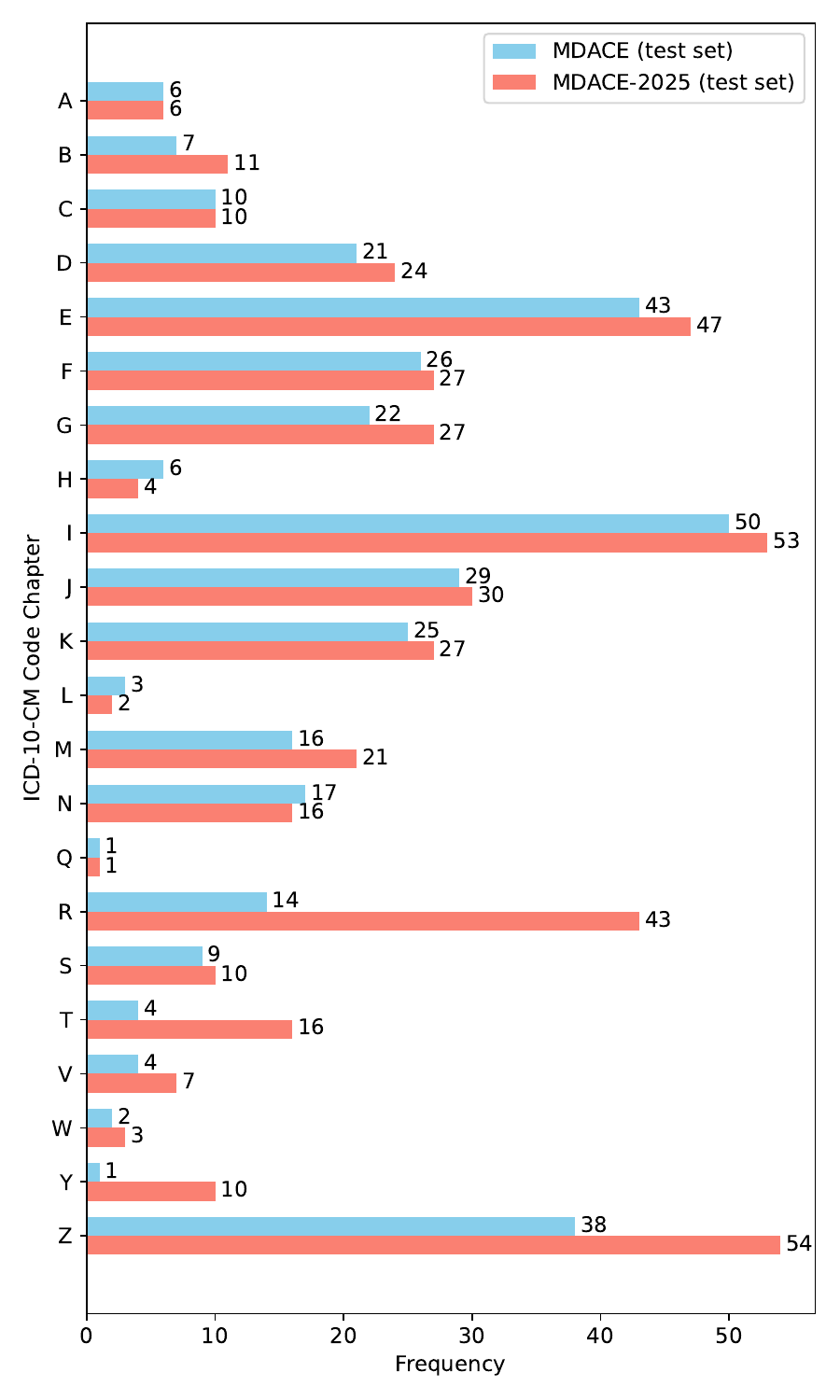}
    \caption{ICD-10-CM code distribution by chapter.}
    \label{fig:code_distribution_between_datasets}
\end{figure}

The test split shared by MDACE and MDACE-2025 contains 61 encounters and 116 notes. We quantify code annotation shifts using retention and addition rates (\S\ref{sec:mdace2025}): overall, 527/678 (77.73\%) of MDACE's codes are retained in MDACE-2025, which adds 378/905 (41.77\%) new codes; for diagnosis (ICD-10-CM) codes, 510/616 (82.79\%) are retained and 376/886 (42.44\%) are added; for procedure (ICD-10-PCS) codes, 17/62 (27.42\%) are retained and 2/19 (10.53\%) are added. Discrepancies arise primarily from guideline updates, additions of supplementary Z codes (e.g., for medical history and allergies), and stricter procedure code restrictions to those supported by operative notes. Figure~\ref{fig:code_distribution_between_datasets} shows the distribution of ICD-10-CM codes by chapter (i.e.\ first character) in the test sets of MDACE and MDACE-2025; Table~\ref{tab:stats_diff_between_datasets} presents an overview of the statistical differences between them.

\begin{table}[t!]
\centering
\scriptsize
\setlength{\tabcolsep}{4pt}
\begin{tabular}{lcc}
\toprule
& MDACE & MDACE-2025 \\
\midrule
\# of ICD-10-CM codes & 616 & 886 \\
\# of unique ICD-10-CM codes & 320 & 451 \\
\# of ICD-10-PCS codes & 62 & 19 \\
\# of unique ICD-10-PCS codes & 43 & 16 \\
\# of avg. codes per encounter & 11.11 & 14.84 \\
\% of encounters w/ justifications & - & 11.48 \\
\bottomrule
\end{tabular}
\caption{
Dataset statistics on the test split shared by MDACE and MDACE-2025. Key shifts include expanded diagnosis annotations, a reduction in procedure annotations attributable to guideline changes, and the addition of expert justification comments for complex coding decisions.
}
\label{tab:stats_diff_between_datasets}
\end{table}

\section{LLM Checkpoints and Coding Guideline Versions} \label{appx:llm_checkpoints_and_guideline_versions}
The LLM checkpoints used in this study are: deepseek-V3.2, qwen3-235b-a22b-instruct-2507, gpt-4o-2024-05-13, gpt-4.1-2025-04-14, and gpt-5-2025-08-07. We empirically set the temperature to 0.2 for all models except GPT-5, which only supports the default temperature (1.0) via the API.

For experiments on the MDACE dataset, we use the April 1st, 2022 coding guidelines and tabular list; this version is chosen over the original 2021 materials because the latter contain an incomplete embedded table of contents. For the MDACE-2025 dataset, we use the April 1st, 2025 version of the guidelines and tabular list.

\section{Description of Baselines} \label{appx:description_of_baselines}
RAG-Coding is compared against these baselines:
\begin{itemize}
    \item \textbf{Tree Search}~\citep{boyle2023automated} introduces an LLM-based tree search algorithm, navigating the ICD ontology recursively to predict codes level by level.
    \item \textbf{MedCodER}~\citep{baksi-etal-2025-medcoder} uses an LLM to extract diagnoses and generate ICD codes. It retrieves additional codes via semantic search on the extracted diagnoses and indexed code descriptions, then re-ranks the combined set of generated and retrieved codes.
    \item \textbf{CLH}~\citep{motzfeldt-etal-2025-code} is an agentic method simulating human coding workflows. It identifies codable text spans, retrieves candidate codes via semantic search against the alphabetical index\footnote{The alphabetical index lists terms alphabetically to locate ICD-10-CM codes. Multiple terms describing the same condition may map to one code.}, and validates them using coding guidelines and the tabular list.
    \item \textbf{PLM-ICD}~\citep{huang-etal-2022-plm} is not LLM-based and is the SOTA method for automated coding~\citep{edin2023automated, gan-etal-2025-aligning, motzfeldt-etal-2025-code}. PLM-ICD uses a BERT-based encoder~\citep{devlin-etal-2019-bert} pre-trained on biomedical texts and is fine-tuned with label-wise attention.
\end{itemize}

Tree Search has an additional parameter defining the maximum number of iterations; we set it to 50, the optimal value reported in the original paper. Similarly, MedCodER has an additional parameter defining the number of retrieved codes per diagnosis; we set it to 1, the optimal value reported in the original paper. PLM-ICD is trained on ICD-10-CM codes from the entire MIMIC-IV dataset and the MDACE training set, where the number of unique codes is approx.\ 16K. This means PLM-ICD covers only about 20\% of the ICD-10-CM code space during inference, while LLM-based methods use the entire space ($\approx$70K).

Dynamic few-shot retrieval (Appendix~\ref{appx:few_shot}) is not applied to any baseline: Tree Search's recursive, level-by-level traversal has no analogous slot for in-context examples, and MedCodER and CLH generate candidates via NER and semantic search over an index rather than LLM free-generation, so injecting retrieved examples would require adding an LLM-generation stage not present in either method.

\subsection{CLH vs.\ RAG-Coding: Representing Coding Knowledge} \label{appx:clh_vs_ragcoding}
CLH~\citep{motzfeldt-etal-2025-code} and RAG-Coding both draw on the ICD-10-CM tabular list and coding guidelines, but differ fundamentally in \emph{how} this knowledge is represented and retrieved. We elaborate on these differences below and illustrate them with a worked example.

CLH retrieves the tabular list as flat, plain-text spans, which creates two practical issues. First, when candidate codes are siblings sharing a parent in the hierarchy (e.g., C41.0 and C41.1 under C41), their instructional notes substantially overlap; feeding duplicated text for each candidate into the LLM pollutes the context, wastes tokens, and slows inference. Second, plain text is difficult to customise: many hospitals maintain their own coding policies, and adding or removing such rules for specific codes means editing unstructured chunks of prose rather than well-defined edges of a graph. Plain-text guideline retrieval poses a related problem. Chapter-specific guidelines span multiple (sometimes 10+) pages, so validating a single code becomes a needle-in-a-haystack search: only one or two sentences may be relevant, while the rest describes rules for unrelated sibling or related codes.

RAG-Coding addresses both issues by representing the tabular list as a KG of \texttt{[subject, predicate, object]} triplets and by summarising guidelines into short, code-specific bullet points. Table~\ref{tab:clh_vs_ragcoding_example} compares both models' knowledge representation for the code E66.09 (other obesity due to excess calories).

\begin{table*}[t]
\centering
\footnotesize
\begin{tabular}{@{}l p{0.42\linewidth} p{0.42\linewidth}@{}}
\toprule
Model & Tabular List & Guidelines \\
\midrule
CLH &
\begin{minipage}[t]{\linewidth}
\vspace{2pt}
\textbf{E66 Overweight and obesity}
\begin{itemize}[noitemsep,topsep=2pt,leftmargin=*]
    \item Code first: obesity complicating pregnancy, childbirth and the puerperium, if applicable (O99.21-)
    \item Use additional code: to identify body mass index (BMI), if known (Z68.-)
    \item Excludes1: adiposogenital dystrophy (E23.6)
\end{itemize}
\dots
\vspace{2pt}
\end{minipage}
&
\begin{minipage}[t]{\linewidth}
\vspace{2pt}
\textbf{Chapter 4: Endocrine, Nutritional, and Metabolic Diseases (E00--E89)}
\begin{itemize}[noitemsep,topsep=2pt,leftmargin=*]
    \item a. Diabetes mellitus -- the diabetes mellitus codes are combination codes that include the type of diabetes mellitus\dots
    \begin{itemize}[noitemsep,leftmargin=*]
        \item Type of diabetes\dots
        \item Type of diabetes mellitus not documented\dots
    \end{itemize}
\end{itemize}
\emph{(chapter continues for several pages, mostly unrelated to E66.09)}
\vspace{2pt}
\end{minipage}
\\
\midrule
RAG-Coding &
\begin{minipage}[t]{\linewidth}
\vspace{2pt}
\texttt{[E66, description, ``overweight and obesity'']}\\
\texttt{[E66, code\_first, O99.21-]}\\
\texttt{[E66, use\_additional\_code, Z68.-]}\\
\texttt{[E66, excludes1, E23.6]}\\
\texttt{[E66.09, ancestor, E66]}\\
\dots
\end{minipage}
&
\begin{minipage}[t]{\linewidth}
\vspace{2pt}
\begin{itemize}[noitemsep,leftmargin=*]
    \item BMI (Z68.-) may be coded with obesity when documented; BMI codes are secondary only.
    \item BMI codes are not assigned during pregnancy.
\end{itemize}
\vspace{2pt}
\end{minipage}
\\
\bottomrule
\end{tabular}
\caption{
Comparison between CLH and RAG-Coding for code E66.09 (other obesity due to excess calories). CLH retrieves the tabular list entry and chapter guideline as plain text, including information irrelevant to E66.09 (e.g., diabetes mellitus rules under the same chapter). RAG-Coding represents the tabular list as KG triplets and summarises guidelines into short, code-specific bullet points.
}
\label{tab:clh_vs_ragcoding_example}
\end{table*}

\section{Code Auditing with LLMs' Internal Knowledge} \label{appx:self_correction}
The ablation results in Table~\ref{tab:ablation_closed_book} show that LLMs cannot effectively correct coding errors using only their internal knowledge. With GPT-4.1, we implemented a prompt-based self-correction mechanism, representing a na\"ive version of SELF-REFINE~\citep{madaan2023self} without external feedback. When given code predictions from the \textbf{\textit{Candidate Generator}} (step 1), the LLM failed to detect false positives and false negatives reliably.

\begin{table}[t!]
\centering
\footnotesize
\begin{tabular}{l*{3}{c}}
\toprule
 & \multicolumn{3}{c}{Micro} \\
\cmidrule(lr){2-4}
& Precision & Recall & F1 \\
\midrule
RAG-Coding (step 1) & 0.34 & \textbf{0.54} & \textbf{0.41} \\
\midrule
with SC & \textbf{0.35} & 0.53 & \textbf{0.41} \\
with SC $\times 2$ & \textbf{0.35} & 0.53 & \textbf{0.41} \\
\bottomrule
\end{tabular}
\caption{
Ablation study on the \emph{MDACE} test set. We use GPT-4.1 as the backbone LLM. `SC' denotes self-correction, where the LLM validates the candidate codes (output of step 1) using only its internal knowledge. `SC $\times$ 2' applies self-correction twice.
}
\label{tab:ablation_closed_book}
\end{table}

\section{Detailed Analysis of Individual Components} \label{appx:detailed_analysis}
This appendix gives the full per-component analysis summarised in \S\ref{sec:ablation_study}, based on a manual examination of intermediate outputs for 20 randomly selected test samples (GPT-4.1 backbone).

\subsection{Candidate Generator (Step 1)}
\textbf{\textit{Candidate Generator}} produces an average of 8.6 candidate codes per sample against 8.0 gold diagnosis codes, achieving a precision of 0.27 and recall of 0.41 on this subset of 20 samples. False positives are heavily skewed by code category. Symptom and sign codes (Chapter R) have a 96\% false positive rate (23 false positives, 1 true positive): these symptoms and signs are documented but should not be assigned because a specific diagnosis is already present. Injury codes (Chapters~S--T) show an 89\% false positive rate (17 false positives, 2 true positives), with errors mainly occurring at the fourth-digit fine-grained level. Recall errors are most prevalent in Z-codes (14 missed), which encode administrative context such as personal and family history, and in circulatory codes (11 missed). Of the 25 Z/circulatory FNs, ``never coded at all'' accounts for 40\%: mostly long-term medications (e.g., coumadin, aspirin) the note mentions but the model never flags as a status code; ``right problem, wrong specificity'' accounts for 20\%: the model found the issue but coded it imprecisely (e.g., downgrading a named subtype to ``unspecified''). Beyond these two error categories, the remaining FNs fall into much smaller categories (e.g., history mistaken for active disease) that are not statistically conclusive.

\subsection{KG-Based Auditor (Step 2)}
\textbf{\textit{KG-Based Auditor}} acts as a conservative pruner: it removes an average of 2.2 codes per sample (from 8.6 to 6.4) and adds zero new codes across all 20 samples. Of the 43 total removals, 39 (91\%) are correct, and 4 (9\%) are incorrect removals of true target codes.

\begin{table}[t!]
\centering
\footnotesize
\begin{tabular}{ll}
\toprule
Reason for Code Removal & Accuracy \\
\midrule
Symptom subsumed by diagnosis           & 13/13 (100\%) \\
Wrong sub-code (laterality/specificity) & 13/13 (100\%) \\
Code description--note mismatch         &  8/8  (100\%) \\
Undocumented/inferred diagnosis         &  5/5  (100\%) \\
\midrule
Comorbidity over-pruning                &  0/4  (0\%)   \\
\midrule
\textbf{Total}                          & 39/43 (91\%)  \\
\bottomrule
\end{tabular}
\caption{Step 2 code removals categorised by reason. Accuracy is defined as the proportion of correct removals (i.e., the removed code was not a true target code).}
\label{tab:step2_removal_categories}
\end{table}

The first four categories in Table~\ref{tab:step2_removal_categories} reflect four KG-driven correction mechanisms:
\begin{itemize}
    \item \textbf{Symptom subsumed by diagnosis:} symptom and sign codes (e.g., R55~\emph{Syncope}, R57.9~\emph{Shock, unspecified}) are correctly removed when hierarchical KG edges reveal that a co-assigned diagnosis already subsumes them.
    \item \textbf{Wrong sub-code (laterality/specificity):} inclusion-term and ancestor edges reveal laterality or sub-type errors (e.g., a lower-extremity arterial thrombosis code selected when the upper extremity is documented).
    \item \textbf{Code description--note mismatch:} the KG code description directly contradicts the medical note, unambiguously flagging a wrong code (e.g., A41.53~\emph{Sepsis due to Serratia} when the note documents Klebsiella).
    \item \textbf{Undocumented/inferred diagnosis:} codes inferred from lab values or imaging alone (e.g., D64.9~\emph{Anaemia} from haemoglobin values; J18.9~\emph{Pneumonia} from a CT infiltrate) are correctly removed because the code descriptions and their KG synonyms are not documented in the medical note.
\end{itemize}

The main systematic failure is \emph{comorbidity over-pruning}: the four incorrect removals involve valid but briefly documented comorbidities: F17.200~\emph{Nicotine dependence}, E11.9~\emph{Type~2 diabetes}, F32.A~\emph{Depression}, and M19.90~\emph{Osteoarthritis}. The KG provides no signal against these codes, yet \textbf{\textit{KG-Based Auditor}} removes them. We hypothesise this is due to insufficient contextual density: each comorbidity appears only once or twice in the note (typically in the past-medical-history section), which becomes harder to keep track of when the medical note is lengthy.

\subsection{Guideline Summariser (Step 3)}
\textbf{\textit{Guideline Summariser}} retrieves applicable guidelines for 90\% of step 2 output codes (115 of 128). The 10\% gap is concentrated in circulatory and respiratory codes, where relevant rules are defined in chapter-level narrative rather than code-specific narrative. When no guideline is found, the code is retained by default in step 4.

\subsection{Guideline-Based Auditor (Step 4)}
\textbf{\textit{Guideline-Based Auditor}} makes 31 changes across the 20 samples (19 removals, 10 additions, and 2 replacements).

\paragraph{Correct Changes.}
Removals are the most reliable action (89\%; 17 of 19 correct): guidelines consistently lead to removal of codes unsupported by explicit documentation or made redundant by a co-assigned code. The two correct replacements represent the \textbf{\textit{Guideline-Based Auditor}}'s highest-value contribution. In one sample, the guideline for I10 prohibits its use when chronic kidney disease is co-documented; the auditor replaces I10~\emph{Hypertension} with I12.9~\emph{Hypertensive chronic kidney disease}, the target code. In another sample, I48.91~(\emph{Atrial fibrillation, unspecified}) is correctly replaced by I48.0~(\emph{Paroxysmal atrial fibrillation}). There are no incorrect replacements. The two correct additions follow a required companion-code pattern: when severe sepsis is documented, coding guidelines mandate both an underlying sepsis code and a severity code; accordingly, the auditor correctly adds A41.9~\emph{Sepsis, unspecified} and R65.21~\emph{Severe sepsis with septic shock}.

\paragraph{Incorrect Changes.}
Addition accuracy is poor (20\%: 2 of 10 correct), revealing two failure modes. The more prevalent is \emph{specificity over-correction} (6 cases): the auditor misinterprets ``prefer specificity when documentation supports it'', assigning a more specific code even when the documentation is insufficient. The second failure mode is \emph{unsupported code insertion} (2 cases): codes added by clinical inference without direct guideline support (e.g., Q60.0~\emph{Renal agenesis} added when only a horseshoe kidney variant was documented). Notably, all incorrect changes in step 4 occur despite the relevant code having a guideline available, confirming that the failure is one of guideline misapplication rather than absence.

\section{Cost Analysis}
RAG-Coding is designed with affordability in mind. We report token consumption and actual costs for RAG-Coding (GPT-4.1 backbone) in Table~\ref{tab:cost}. On average, RAG-Coding costs \$0.25 USD to fully code an encounter in the MDACE validation set, where each encounter contains an average of 2.03 notes. Step 3, which involves retrieving and summarising coding guidelines, is the most expensive component, incurring 68\% of the overall cost. However, these summaries can be cached and reused, so the cost of step 3 approaches zero over time, reducing the long-term inference cost to approximately \$0.08 per encounter. Consequently, RAG-Coding remains practical and cost-effective even on large-scale datasets. For context, manual medical coding costs an estimated \$9.66--13.69 per encounter, based on a median US medical coder wage of \$24.16/hour~\citep{bls2025medicalrecords} and a reported 24--34 minutes to code a single inpatient encounter~\citep{kim2021read, gao2024optimising}. RAG-Coding's per-encounter cost (\$0.25, dropping to \$0.08 as the guideline-summary cache fills) is therefore much lower than manual coding cost.

\begin{table}[t!]
\centering
\footnotesize
\begin{tabular}{lrc}
\toprule
& \begin{tabular}{@{}c@{}}Tokens \\ Consumed\end{tabular} & \begin{tabular}{@{}c@{}}API Cost\\(\$ USD)\end{tabular} \\
\midrule
RAG-Coding (step 1) & 13,046.30 & 0.03 \\
RAG-Coding (step 2) & 6,779.02 & 0.02 \\
RAG-Coding (step 3) & 85,636.73 & 0.17 \\
RAG-Coding (step 4) & 7,360.03 & 0.03 \\
\midrule
Total & 112,822.08 & 0.25 \\
\bottomrule
\end{tabular}
\caption{
Cost breakdown for RAG-Coding (GPT-4.1 backbone). This experiment was conducted on the \emph{MDACE} validation set. Each row shows token (prompt and completion) usage and actual API cost for individual RAG-Coding components. The final row is the total across all components. Note: Step 3 output is cached, so its cost approaches zero in repeated/long-term use.
}
\label{tab:cost}
\end{table}

\section{Full Results} \label{appx:full_results}
The full versions of the main experiment results on the MDACE and MDACE-2025 test sets are presented in Table~\ref{tab:full_results_against_llm_based_baselines},  Table~\ref{tab:full_results_against_plm_icd}, Table~\ref{tab:mdace_2025_full_results_against_llm_baselines}, and Table~\ref{tab:mdace_2025_full_results_against_plm_icd}.

\begin{table*}[t!]
\centering
\scriptsize
\begin{tabular}{l|l*{6}{c}}
\toprule
\multicolumn{1}{l}{LLM} & Method & \multicolumn{3}{c}{Micro} & \multicolumn{3}{c}{Macro} \\
\cmidrule(lr){3-5}
\cmidrule(lr){6-8}
\multicolumn{1}{l}{} & & Precision & Recall & F1 & Precision & Recall & F1 \\
\midrule
\multirow{4}{*}{Qwen3}
& Tree Search & 0.08 $\pm$ 0.01 & 0.17 $\pm$ 0.00 & 0.11 $\pm$ 0.01 & 0.05 $\pm$ 0.01 & 0.05 $\pm$ 0.00 & 0.04 $\pm$ 0.01 \\
& MedCodER & 0.29 $\pm$ 0.02 & 0.36 $\pm$ 0.03 & 0.32 $\pm$ 0.02 & 0.16 $\pm$ 0.01 & 0.16 $\pm$ 0.01 & 0.15 $\pm$ 0.01 \\
& CLH & 0.27 $\pm$ 0.03 & \textbf{0.49 $\pm$ 0.03} & 0.35 $\pm$ 0.03 & 0.17 $\pm$ 0.02 & \textbf{0.18 $\pm$ 0.01} & 0.17 $\pm$ 0.01 \\
& RAG-Coding & \textbf{0.43 $\pm$ 0.02} & 0.39 $\pm$ 0.01 & \textbf{0.41 $\pm$ 0.02} & \textbf{0.20 $\pm$ 0.01} & \textbf{0.18 $\pm$ 0.01} & \textbf{0.18 $\pm$ 0.01} \\
\midrule
\multirow{4}{*}{Deepseek-V3.2}
& Tree Search & 0.12 $\pm$ 0.01 & 0.25 $\pm$ 0.02 & 0.16 $\pm$ 0.02 & 0.07 $\pm$ 0.01 & 0.08 $\pm$ 0.01 & 0.07 $\pm$ 0.01 \\
& MedCodER & 0.33 $\pm$ 0.02 & \textbf{0.48 $\pm$ 0.02} & 0.39 $\pm$ 0.02 & 0.18 $\pm$ 0.01 & 0.19 $\pm$ 0.00 & 0.17 $\pm$ 0.01 \\
& CLH & 0.37 $\pm$ 0.04 & 0.46 $\pm$ 0.03 & 0.41 $\pm$ 0.03 & 0.21 $\pm$ 0.02 & 0.21 $\pm$ 0.01 & 0.21 $\pm$ 0.02 \\
& RAG-Coding & \textbf{0.46 $\pm$ 0.03} & 0.43 $\pm$ 0.02 & \textbf{0.44 $\pm$ 0.03} & \textbf{0.25 $\pm$ 0.02} & \textbf{0.22 $\pm$ 0.02} & \textbf{0.22 $\pm$ 0.01} \\
\midrule
\multirow{4}{*}{GPT-4o}
& Tree Search & 0.14 $\pm$ 0.01 & 0.26 $\pm$ 0.03 & 0.18 $\pm$ 0.02 & 0.09 $\pm$ 0.01 & 0.09 $\pm$ 0.01 & 0.09 $\pm$ 0.01 \\
& MedCodER & 0.39 $\pm$ 0.01 & 0.36 $\pm$ 0.01 & 0.37 $\pm$ 0.01 & 0.19 $\pm$ 0.01 & 0.18 $\pm$ 0.02 & 0.17 $\pm$ 0.01 \\
& CLH & 0.29 $\pm$ 0.02 & 0.47 $\pm$ 0.00 & 0.36 $\pm$ 0.02 & 0.18 $\pm$ 0.01 & 0.19 $\pm$ 0.01 & 0.18 $\pm$ 0.01 \\
& RAG-Coding & \textbf{0.42 $\pm$ 0.01} & \textbf{0.48 $\pm$ 0.01} & \textbf{0.45 $\pm$ 0.01} & \textbf{0.21 $\pm$ 0.01} & \textbf{0.22 $\pm$ 0.01} & \textbf{0.21 $\pm$ 0.01} \\
\midrule
\multirow{4}{*}{GPT-4.1}
& Tree Search & 0.11 $\pm$ 0.00 & 0.22 $\pm$ 0.01 & 0.15 $\pm$ 0.01 & 0.08 $\pm$ 0.01 & 0.08 $\pm$ 0.01 & 0.07 $\pm$ 0.01 \\
& MedCodER & 0.31 $\pm$ 0.01 & 0.44 $\pm$ 0.01 & 0.36 $\pm$ 0.01 & 0.17 $\pm$ 0.01 & 0.18 $\pm$ 0.01 & 0.17 $\pm$ 0.01 \\
& CLH & 0.32 $\pm$ 0.01 & 0.52 $\pm$ 0.01 & 0.40 $\pm$ 0.00 & 0.20 $\pm$ 0.00 & 0.22 $\pm$ 0.01 & 0.20 $\pm$ 0.01 \\
& RAG-Coding & \textbf{0.47 $\pm$ 0.02} & \textbf{0.59 $\pm$ 0.05} & \textbf{0.53 $\pm$ 0.03} & \textbf{0.27 $\pm$ 0.03} & \textbf{0.29 $\pm$ 0.03} & \textbf{0.27 $\pm$ 0.03} \\
\midrule
\multirow{4}{*}{GPT-5}
& Tree Search & 0.17 $\pm$ 0.01 & 0.33 $\pm$ 0.00 & 0.22 $\pm$ 0.00 & 0.10 $\pm$ 0.00 & 0.11 $\pm$ 0.00 & 0.09 $\pm$ 0.01 \\
& MedCodER & 0.37 $\pm$ 0.01 & 0.42 $\pm$ 0.01 & 0.40 $\pm$ 0.01 & 0.21 $\pm$ 0.01 & 0.22 $\pm$ 0.01 & 0.20 $\pm$ 0.01 \\
& CLH & 0.40 $\pm$ 0.01 & 0.55 $\pm$ 0.02 & 0.46 $\pm$ 0.02 & 0.24 $\pm$ 0.01 & 0.26 $\pm$ 0.01 & 0.24 $\pm$ 0.01 \\
& RAG-Coding & \textbf{0.46 $\pm$ 0.01} & \textbf{0.66 $\pm$ 0.01} & \textbf{0.54 $\pm$ 0.01} & \textbf{0.31 $\pm$ 0.01} & \textbf{0.33 $\pm$ 0.01} & \textbf{0.32 $\pm$ 0.00} \\
\bottomrule
\end{tabular}
\caption{
Extended version of Table~\ref{tab:main_results}'s \#C=70K rows, reporting mean $\pm$ standard deviation over three runs on the \emph{MDACE} test set. All methods use the entire code space during inference ($\approx$ 70K codes).
}
\label{tab:full_results_against_llm_based_baselines}
\end{table*}

\begin{table*}[t!]
\centering
\scriptsize
\begin{tabular}{l|l*{6}{c}}
\toprule
\multicolumn{1}{l}{LLM} & Method & \multicolumn{3}{c}{Micro} & \multicolumn{3}{c}{Macro} \\
\cmidrule(lr){3-5}
\cmidrule(lr){6-8}
\multicolumn{1}{l}{} & & Precision & Recall & F1 & Precision & Recall & F1 \\
\midrule
\quad- & PLM-ICD & \textbf{0.54 $\pm$ 0.02} & 0.53 $\pm$ 0.02 & 0.53 $\pm$ 0.02 & 0.29 $\pm$ 0.02 & 0.30 $\pm$ 0.01 & 0.30 $\pm$ 0.02 \\
\midrule
\multirow{4}{*}{Qwen3}
& Tree Search & 0.09 $\pm$ 0.01 & 0.17 $\pm$ 0.01 & 0.12 $\pm$ 0.01 & 0.05 $\pm$ 0.00 & 0.05 $\pm$ 0.00 & 0.05 $\pm$ 0.00 \\
& MedCodER & 0.32 $\pm$ 0.02 & 0.36 $\pm$ 0.02 & 0.34 $\pm$ 0.02 & 0.18 $\pm$ 0.01 & 0.18 $\pm$ 0.02 & 0.17 $\pm$ 0.02 \\
& CLH & 0.29 $\pm$ 0.02 & 0.49 $\pm$ 0.02 & 0.36 $\pm$ 0.02 & 0.19 $\pm$ 0.02 & 0.20 $\pm$ 0.01 & 0.18 $\pm$ 0.02 \\
& RAG-Coding & 0.47 $\pm$ 0.02 & 0.39 $\pm$ 0.02 & 0.43 $\pm$ 0.02 & 0.21 $\pm$ 0.01 & 0.20 $\pm$ 0.01 & 0.19 $\pm$ 0.02 \\
\midrule
\multirow{4}{*}{Deepseek-V3.2}
& Tree Search & 0.13 $\pm$ 0.01 & 0.25 $\pm$ 0.00 & 0.17 $\pm$ 0.01 & 0.08 $\pm$ 0.00 & 0.09 $\pm$ 0.00 & 0.08 $\pm$ 0.00 \\
& MedCodER & 0.34 $\pm$ 0.03 & 0.48 $\pm$ 0.02 & 0.40 $\pm$ 0.02 & 0.19 $\pm$ 0.02 & 0.21 $\pm$ 0.01 & 0.19 $\pm$ 0.02 \\
& CLH & 0.40 $\pm$ 0.02 & 0.46 $\pm$ 0.02 & 0.43 $\pm$ 0.02 & 0.23 $\pm$ 0.01 & 0.25 $\pm$ 0.01 & 0.23 $\pm$ 0.01 \\
& RAG-Coding & 0.50 $\pm$ 0.03 & 0.43 $\pm$ 0.03 & 0.46 $\pm$ 0.03 & 0.27 $\pm$ 0.02 & 0.24 $\pm$ 0.01 & 0.24 $\pm$ 0.01 \\
\midrule
\multirow{4}{*}{GPT-4o}
& Tree Search & 0.16 $\pm$ 0.00 & 0.26 $\pm$ 0.01 & 0.21 $\pm$ 0.01 & 0.10 $\pm$ 0.00 & 0.11 $\pm$ 0.00 & 0.10 $\pm$ 0.00 \\
& MedCodER & 0.41 $\pm$ 0.00 & 0.36 $\pm$ 0.00 & 0.39 $\pm$ 0.01 & 0.20 $\pm$ 0.00 & 0.19 $\pm$ 0.00 & 0.19 $\pm$ 0.00 \\
& CLH & 0.32 $\pm$ 0.02 & 0.47 $\pm$ 0.00 & 0.38 $\pm$ 0.01 & 0.20 $\pm$ 0.00 & 0.21 $\pm$ 0.01 & 0.20 $\pm$ 0.01 \\
& RAG-Coding & 0.44 $\pm$ 0.01 & 0.48 $\pm$ 0.01 & 0.46 $\pm$ 0.01 & 0.21 $\pm$ 0.01 & 0.22 $\pm$ 0.01 & 0.21 $\pm$ 0.01 \\
\midrule
\multirow{4}{*}{GPT-4.1}
& Tree Search & 0.14 $\pm$ 0.00 & 0.22 $\pm$ 0.01 & 0.17 $\pm$ 0.00 & 0.09 $\pm$ 0.00 & 0.09 $\pm$ 0.00 & 0.08 $\pm$ 0.00 \\
& MedCodER & 0.33 $\pm$ 0.00 & 0.43 $\pm$ 0.01 & 0.38 $\pm$ 0.01 & 0.17 $\pm$ 0.01 & 0.20 $\pm$ 0.01 & 0.17 $\pm$ 0.01 \\
& CLH & 0.35 $\pm$ 0.01 & 0.52 $\pm$ 0.01 & 0.42 $\pm$ 0.00 & 0.22 $\pm$ 0.00 & 0.23 $\pm$ 0.01 & 0.21 $\pm$ 0.01 \\
& RAG-Coding & 0.51 $\pm$ 0.04 & 0.53 $\pm$ 0.01 & 0.52 $\pm$ 0.02 & 0.26 $\pm$ 0.01 & 0.27 $\pm$ 0.01 & 0.25 $\pm$ 0.01 \\
\midrule
\multirow{4}{*}{GPT-5}
& Tree Search & 0.19 $\pm$ 0.00 & 0.33 $\pm$ 0.00 & 0.24 $\pm$ 0.00 & 0.11 $\pm$ 0.00 & 0.12 $\pm$ 0.00 & 0.10 $\pm$ 0.00 \\
& MedCodER & 0.39 $\pm$ 0.00 & 0.42 $\pm$ 0.01 & 0.41 $\pm$ 0.01 & 0.23 $\pm$ 0.01 & 0.23 $\pm$ 0.01 & 0.21 $\pm$ 0.01 \\
& CLH & 0.43 $\pm$ 0.01 & 0.55 $\pm$ 0.02 & 0.48 $\pm$ 0.01 & 0.26 $\pm$ 0.01 & 0.29 $\pm$ 0.01 & 0.27 $\pm$ 0.01 \\
& RAG-Coding & 0.48 $\pm$ 0.03 & \textbf{0.64 $\pm$ 0.03} & \textbf{0.55 $\pm$ 0.01} & \textbf{0.30 $\pm$ 0.02} & \textbf{0.32 $\pm$ 0.02} & \textbf{0.31 $\pm$ 0.02} \\
\bottomrule
\end{tabular}
\caption{
Extended version of Table~\ref{tab:main_results}'s \#C=16K rows, reporting mean $\pm$ standard deviation over three runs on the \emph{MDACE} test set. All LLM-based methods apply a post-hoc filter that removes any codes outside the 16K space used by PLM-ICD, enabling a fairer comparison. Unlike Table~\ref{tab:full_results_against_llm_based_baselines}, bold here denotes the best result overall (i.e., across all LLM-based methods and PLM-ICD), not the best result for each LLM backbone.
}
\label{tab:full_results_against_plm_icd}
\end{table*}

\begin{table*}[t!]
\centering
\scriptsize
\begin{tabular}{l|l*{6}{c}}
\toprule
\multicolumn{1}{l}{LLM} & Method & \multicolumn{3}{c}{Micro} & \multicolumn{3}{c}{Macro} \\
\cmidrule(lr){3-5}
\cmidrule(lr){6-8}
\multicolumn{1}{l}{} & & Precision & Recall & F1 & Precision & Recall & F1 \\
\midrule
\multirow{4}{*}{Qwen3} 
& Tree Search & 0.10 $\pm$ 0.01 & 0.15 $\pm$ 0.02 & 0.12 $\pm$ 0.01 & 0.06 $\pm$ 0.00 & 0.06 $\pm$ 0.00 & 0.05 $\pm$ 0.00 \\
& MedCodER & 0.32 $\pm$ 0.02 & 0.27 $\pm$ 0.02 & 0.29 $\pm$ 0.01 & 0.17 $\pm$ 0.01 & 0.15 $\pm$ 0.01 & 0.15 $\pm$ 0.01 \\
& CLH & 0.30 $\pm$ 0.02 & \textbf{0.38 $\pm$ 0.02} & 0.33 $\pm$ 0.02 & 0.18 $\pm$ 0.02 & \textbf{0.18 $\pm$ 0.01} & \textbf{0.17 $\pm$ 0.01} \\
& RAG-Coding & \textbf{0.46 $\pm$ 0.03} & 0.29 $\pm$ 0.02 & \textbf{0.35 $\pm$ 0.02} & \textbf{0.19 $\pm$ 0.02} & 0.17 $\pm$ 0.01 & \textbf{0.17 $\pm$ 0.02} \\
\midrule
\multirow{4}{*}{Deepseek-V3.2} 
& Tree Search & 0.14 $\pm$ 0.01 & 0.22 $\pm$ 0.00 & 0.17 $\pm$ 0.00 & 0.09 $\pm$ 0.00 & 0.09 $\pm$ 0.00 & 0.08 $\pm$ 0.00 \\
& MedCodER & 0.34 $\pm$ 0.02 & \textbf{0.35 $\pm$ 0.01} & 0.34 $\pm$ 0.01 & 0.18 $\pm$ 0.01 & 0.18 $\pm$ 0.00 & 0.17 $\pm$ 0.01 \\
& CLH & 0.41 $\pm$ 0.02 & \textbf{0.35 $\pm$ 0.03} & 0.38 $\pm$ 0.02 & 0.20 $\pm$ 0.02 & \textbf{0.21 $\pm$ 0.01} & 0.20 $\pm$ 0.01 \\
& RAG-Coding & \textbf{0.48 $\pm$ 0.03} & \textbf{0.35 $\pm$ 0.02} & \textbf{0.40 $\pm$ 0.02} & \textbf{0.23 $\pm$ 0.02} & 0.20 $\pm$ 0.01 & \textbf{0.21 $\pm$ 0.01} \\
\midrule
\multirow{4}{*}{GPT-4o}
& Tree Search & 0.17 $\pm$ 0.01 & 0.22 $\pm$ 0.02 & 0.19 $\pm$ 0.02 & 0.11 $\pm$ 0.01 & 0.10 $\pm$ 0.01 & 0.10 $\pm$ 0.01 \\
& MedCodER & 0.42 $\pm$ 0.01 & 0.26 $\pm$ 0.01 & 0.32 $\pm$ 0.00 & 0.18 $\pm$ 0.01 & 0.16 $\pm$ 0.01 & 0.16 $\pm$ 0.01 \\
& CLH & 0.32 $\pm$ 0.03 & \textbf{0.37 $\pm$ 0.01} & 0.35 $\pm$ 0.02 & 0.18 $\pm$ 0.01 & \textbf{0.19 $\pm$ 0.01} & 0.18 $\pm$ 0.02 \\
& RAG-Coding & \textbf{0.44 $\pm$ 0.02} & 0.34 $\pm$ 0.01 & \textbf{0.39 $\pm$ 0.02} &\textbf{0.19 $\pm$ 0.01} & \textbf{0.19 $\pm$ 0.01} & \textbf{0.19 $\pm$ 0.01} \\
\midrule
\multirow{4}{*}{GPT-4.1}
& Tree Search & 0.14 $\pm$ 0.01 & 0.19 $\pm$ 0.01 & 0.16 $\pm$ 0.01 & 0.09 $\pm$ 0.01 & 0.09 $\pm$ 0.01 & 0.08 $\pm$ 0.01 \\
& MedCodER & 0.35 $\pm$ 0.01 & 0.33 $\pm$ 0.01 & 0.34 $\pm$ 0.01 & 0.18 $\pm$ 0.00 & 0.18 $\pm$ 0.00 & 0.17 $\pm$ 0.00 \\
& CLH & 0.37 $\pm$ 0.02 & 0.39 $\pm$ 0.01 & 0.38 $\pm$ 0.01 & 0.21 $\pm$ 0.01 & 0.22 $\pm$ 0.01 & 0.21 $\pm$ 0.02 \\
& RAG-Coding & \textbf{0.53 $\pm$ 0.03} & \textbf{0.40 $\pm$ 0.01} & \textbf{0.46 $\pm$ 0.01} & \textbf{0.25 $\pm$ 0.01} & \textbf{0.23 $\pm$ 0.01} & \textbf{0.23 $\pm$ 0.01} \\
\midrule
\multirow{4}{*}{GPT-5}
& Tree Search & 0.20 $\pm$ 0.00 & 0.28 $\pm$ 0.00 & 0.24 $\pm$ 0.01 & 0.12 $\pm$ 0.01 & 0.13 $\pm$ 0.01 & 0.12 $\pm$ 0.00 \\ 
& MedCodER & 0.43 $\pm$ 0.02 & 0.35 $\pm$ 0.01 & 0.38 $\pm$ 0.01 & 0.24 $\pm$ 0.01 & 0.23 $\pm$ 0.01 & 0.23 $\pm$ 0.01 \\
& CLH & 0.45 $\pm$ 0.02 & 0.44 $\pm$ 0.02 & 0.45 $\pm$ 0.02 & 0.26 $\pm$ 0.02 & 0.27 $\pm$ 0.03 & 0.26 $\pm$ 0.03 \\
& RAG-Coding & \textbf{0.57 $\pm$ 0.02} & \textbf{0.55 $\pm$ 0.03} & \textbf{0.56 $\pm$ 0.02} & \textbf{0.34 $\pm$ 0.02} & \textbf{0.34 $\pm$ 0.02} & \textbf{0.33 $\pm$ 0.02} \\
\bottomrule
\end{tabular}
\caption{
Extended version of Table~\ref{tab:results_against_baselines_mdace2025}, covering more LLM backbones and reporting mean $\pm$ standard deviation over three runs on the \emph{MDACE-2025} test set. All methods use the entire code space during inference ($\approx$ 70K codes).
}
\label{tab:mdace_2025_full_results_against_llm_baselines}
\end{table*}

\begin{table*}[t!]
\centering
\scriptsize
\begin{tabular}{l|l*{6}{c}}
\toprule
\multicolumn{1}{l}{LLM} & Method & \multicolumn{3}{c}{Micro} & \multicolumn{3}{c}{Macro} \\
\cmidrule(lr){3-5}
\cmidrule(lr){6-8}
\multicolumn{1}{l}{} & & Precision & Recall & F1 & Precision & Recall & F1 \\
\midrule
- & PLM-ICD & \textbf{0.62 $\pm$ 0.02} & 0.39 $\pm$ 0.01 & 0.48 $\pm$ 0.02 & 0.27 $\pm$ 0.01 & 0.25 $\pm$ 0.01 & 0.26 $\pm$ 0.01 \\
\midrule
\multirow{4}{*}{Qwen3}
& Tree Search & 0.12 $\pm$ 0.01 & 0.15 $\pm$ 0.02 & 0.13 $\pm$ 0.02 & 0.06 $\pm$ 0.01 & 0.06 $\pm$ 0.01 & 0.06 $\pm$ 0.01 \\
& MedCodER & 0.35 $\pm$ 0.02 & 0.27 $\pm$ 0.01 & 0.30 $\pm$ 0.02 & 0.18 $\pm$ 0.01 & 0.16 $\pm$ 0.01 & 0.16 $\pm$ 0.01 \\
& CLH & 0.32 $\pm$ 0.03 & 0.38 $\pm$ 0.02 & 0.35 $\pm$ 0.02 & 0.19 $\pm$ 0.02 & 0.20 $\pm$ 0.02 & 0.18 $\pm$ 0.02 \\
& RAG-Coding & 0.50 $\pm$ 0.02 & 0.29 $\pm$ 0.02 & 0.36 $\pm$ 0.02 & 0.20 $\pm$ 0.01 & 0.17 $\pm$ 0.01 & 0.18 $\pm$ 0.01 \\
\midrule
\multirow{4}{*}{Deepseek-V3.2}
& Tree Search & 0.16 $\pm$ 0.00 & 0.21 $\pm$ 0.00 & 0.18 $\pm$ 0.00 & 0.10 $\pm$ 0.00 & 0.10 $\pm$ 0.00 & 0.09 $\pm$ 0.00 \\
& MedCodER & 0.36 $\pm$ 0.01 & 0.35 $\pm$ 0.01 & 0.35 $\pm$ 0.01 & 0.19 $\pm$ 0.01 & 0.19 $\pm$ 0.01 & 0.18 $\pm$ 0.01 \\
& CLH & 0.44 $\pm$ 0.03 & 0.35 $\pm$ 0.02 & 0.39 $\pm$ 0.03 & 0.22 $\pm$ 0.02 & 0.23 $\pm$ 0.01 & 0.22 $\pm$ 0.01 \\
& RAG-Coding & 0.52 $\pm$ 0.02 & 0.35 $\pm$ 0.02 & 0.42 $\pm$ 0.01 & 0.24 $\pm$ 0.01 & 0.20 $\pm$ 0.01 & 0.22 $\pm$ 0.01 \\
\midrule
\multirow{4}{*}{GPT-4o}
& Tree Search & 0.19 $\pm$ 0.00 & 0.22 $\pm$ 0.00 & 0.21 $\pm$ 0.00 & 0.12 $\pm$ 0.00 & 0.11 $\pm$ 0.01 & 0.11 $\pm$ 0.01 \\
& MedCodER & 0.43 $\pm$ 0.01 & 0.26 $\pm$ 0.01 & 0.33 $\pm$ 0.01 & 0.18 $\pm$ 0.01 & 0.17 $\pm$ 0.00 & 0.17 $\pm$ 0.01 \\
& CLH & 0.34 $\pm$ 0.04 & 0.36 $\pm$ 0.01 & 0.35 $\pm$ 0.02 & 0.20 $\pm$ 0.01 & 0.21 $\pm$ 0.02 & 0.19 $\pm$ 0.01 \\
& RAG-Coding & 0.45 $\pm$ 0.02 & 0.34 $\pm$ 0.01 & 0.39 $\pm$ 0.01 & 0.19 $\pm$ 0.01 & 0.19 $\pm$ 0.01 & 0.18 $\pm$ 0.01 \\
\midrule
\multirow{4}{*}{GPT-4.1}
& Tree Search & 0.17 $\pm$ 0.00 & 0.19 $\pm$ 0.00 & 0.18 $\pm$ 0.00 & 0.10 $\pm$ 0.00 & 0.10 $\pm$ 0.00 & 0.09 $\pm$ 0.00 \\
& MedCodER & 0.36 $\pm$ 0.01 & 0.33 $\pm$ 0.01 & 0.35 $\pm$ 0.01 & 0.19 $\pm$ 0.00 & 0.19 $\pm$ 0.00 & 0.18 $\pm$ 0.00 \\
& CLH & 0.39 $\pm$ 0.01 & 0.39 $\pm$ 0.00 & 0.39 $\pm$ 0.00 & 0.21 $\pm$ 0.01 & 0.23 $\pm$ 0.01 & 0.21 $\pm$ 0.00 \\
& RAG-Coding & 0.56 $\pm$ 0.03 & 0.40 $\pm$ 0.01 & 0.46 $\pm$ 0.01 & 0.25 $\pm$ 0.01 & 0.24 $\pm$ 0.00 & 0.23 $\pm$ 0.01 \\
\midrule
\multirow{4}{*}{GPT-5}
& Tree Search & 0.23 $\pm$ 0.01 & 0.28 $\pm$ 0.00 & 0.25 $\pm$ 0.00 & 0.13 $\pm$ 0.00 & 0.13 $\pm$ 0.00 & 0.11 $\pm$ 0.01 \\
& MedCodER & 0.46 $\pm$ 0.00 & 0.34 $\pm$ 0.01 & 0.40 $\pm$ 0.01 & 0.25 $\pm$ 0.00 & 0.24 $\pm$ 0.01 & 0.23 $\pm$ 0.01 \\
& CLH & 0.49 $\pm$ 0.01 & 0.42 $\pm$ 0.02 & 0.45 $\pm$ 0.01 & 0.27 $\pm$ 0.01 & 0.28 $\pm$ 0.01 & 0.26 $\pm$ 0.01 \\
& RAG-Coding & 0.58 $\pm$ 0.01 & \textbf{0.55 $\pm$ 0.03} & \textbf{0.56 $\pm$ 0.02} & \textbf{0.35 $\pm$ 0.02} & \textbf{0.34 $\pm$ 0.02} & \textbf{0.34 $\pm$ 0.02} \\
\bottomrule
\end{tabular}
\caption{
Extended version of Table~\ref{tab:results_against_baselines_mdace2025}, covering more LLM backbones and reporting mean $\pm$ standard deviation over three runs on the \emph{MDACE-2025} test set. All LLM-based methods apply a post-hoc filter that removes any codes outside the 16K space used by PLM-ICD, enabling a fairer comparison. Unlike Table~\ref{tab:mdace_2025_full_results_against_llm_baselines}, bold here denotes the best result overall (i.e., across all LLM-based methods and PLM-ICD), not the best result for each LLM backbone.
}
\label{tab:mdace_2025_full_results_against_plm_icd}
\end{table*}

\section{Potential Risks}
While our proposed RAG-Coding method enhances medical coding performance by grounding predictions in retrieved clinical references, it can produce incorrect code suggestions that could affect financial and research outcomes if used without review. Consistent with prior work~\citep{gan-etal-2025-aligning, motzfeldt-etal-2025-code}, we consider fully automated coding not yet ready to replace professional coders, and RAG-Coding is best used as an assistive tool whose suggestions are reviewed by professional coders. The introduced MDACE-2025 dataset is a re-annotated version of public datasets and therefore poses no new privacy risks.

\end{document}